\documentclass[runningheads]{llncs}

\usepackage{eccv}

\usepackage{eccvabbrv}

\usepackage{graphicx}
\usepackage{booktabs}

\usepackage{xcolor} 
\usepackage{paralist}
\usepackage[ruled,vlined]{algorithm2e}
\usepackage{multirow}
\usepackage{multicol}
\usepackage{array}
\usepackage{booktabs}
\usepackage[table]{xcolor}
\usepackage{amsmath}
\usepackage{makecell}
\usepackage{pifont}

\newcommand{\cmark}{\textcolor{green!50!black}{\ding{51}}} 
\newcommand{\xmark}{\textcolor{red!60!black}{\ding{55}}}   
\newcommand{\oursrow}{\rowcolor{teal!10}\cellcolor{white}}

\usepackage[accsupp]{axessibility}  

\usepackage{hyperref}

\usepackage{orcidlink}

\begin{document}

\title{{TaskTok}: Delving into Task Tokens for Task-driven Image Restoration} 


\author{Hongjae Lee\orcidlink{0000-0001-5312-139X} \and
Sojung Kang\orcidlink{0009-0009-6928-2900} \and
Jaeseong Yu\orcidlink{0009-0005-1427-6651} \and
Seung-Won Jung\thanks{Corresponding author.}\orcidlink{0000-0002-0319-4467}}

\authorrunning{H. Lee \etal}

\institute{Korea University\\
\email{\{jimmy9704, topazsco, jsyu624, swjung83\}@korea.ac.kr}}

\maketitle

\begin{figure}[h!]
  \centering
  \includegraphics[width=.97\textwidth]{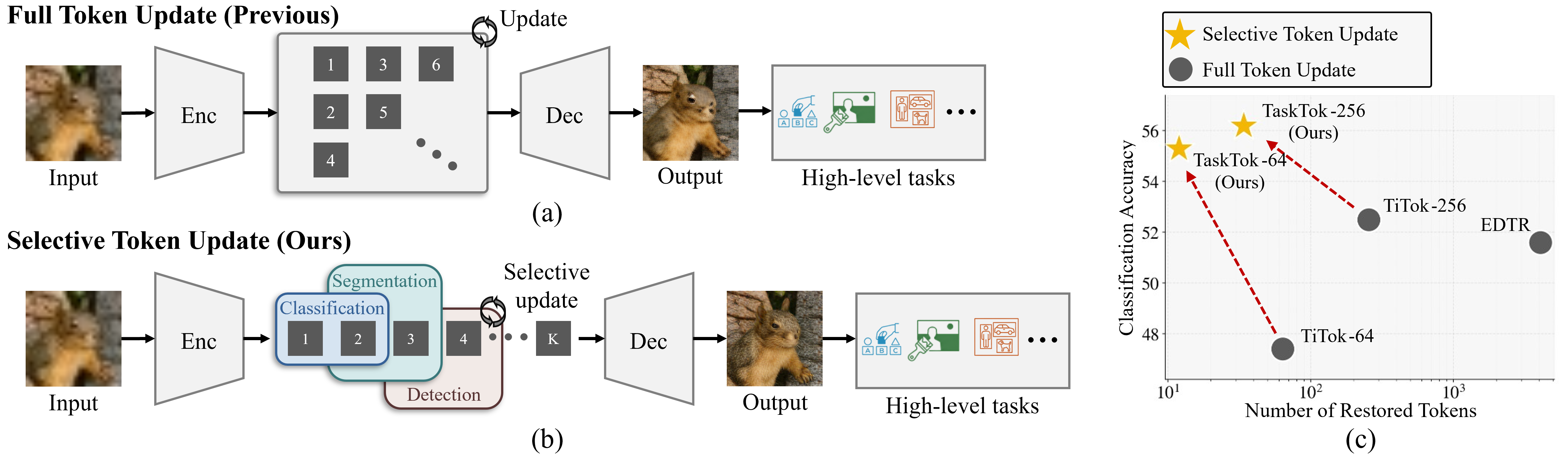}
  \caption{Overview of \textbf{TaskTok}, a task-driven image restoration framework. (a) Existing generative prior-based methods update all latent tokens. (b) In contrast, TaskTok selectively restores only task-relevant tokens, improving performance on high-level tasks. (c) Classification accuracy vs. number of restored tokens. TaskTok achieves higher accuracy while updating substantially fewer tokens than full-token update methods.
  }
  \label{fig:fig1}
\end{figure}

\vspace{-1.8em}

\begin{abstract}
While traditional image restoration focuses on perceptual quality, Task-Driven Image Restoration (TDIR) aims to maximize the performance of downstream high-level vision tasks. Recent approaches leveraging generative priors have shown promise for TDIR; however, they typically suffer from computational inefficiency and potential semantic alteration by indiscriminately updating all latent tokens. In this paper, we posit that not all visual information is equally important for machine perception. Through an analysis of the latent token space, we observe that task-relevant cues are unevenly distributed across the token sequence, exhibiting index-wise specialization. This suggests that selectively refining a subset of tokens can be sufficient for task-driven objectives. Leveraging this insight, we propose TaskTok, a novel framework that selectively restores only task-relevant tokens via a learnable token switch and a lightweight token refinement module. Extensive experiments across image classification, semantic segmentation, and object detection demonstrate that TaskTok significantly enhances task performance with high computational efficiency. The source code is available at our project page \url{https://github.com/jimmy9704/TaskTok}.
\end{abstract}

\section{Introduction}
\label{sec:intro}

\begin{figure}[t!]
  \centering
  \includegraphics[width=\textwidth]{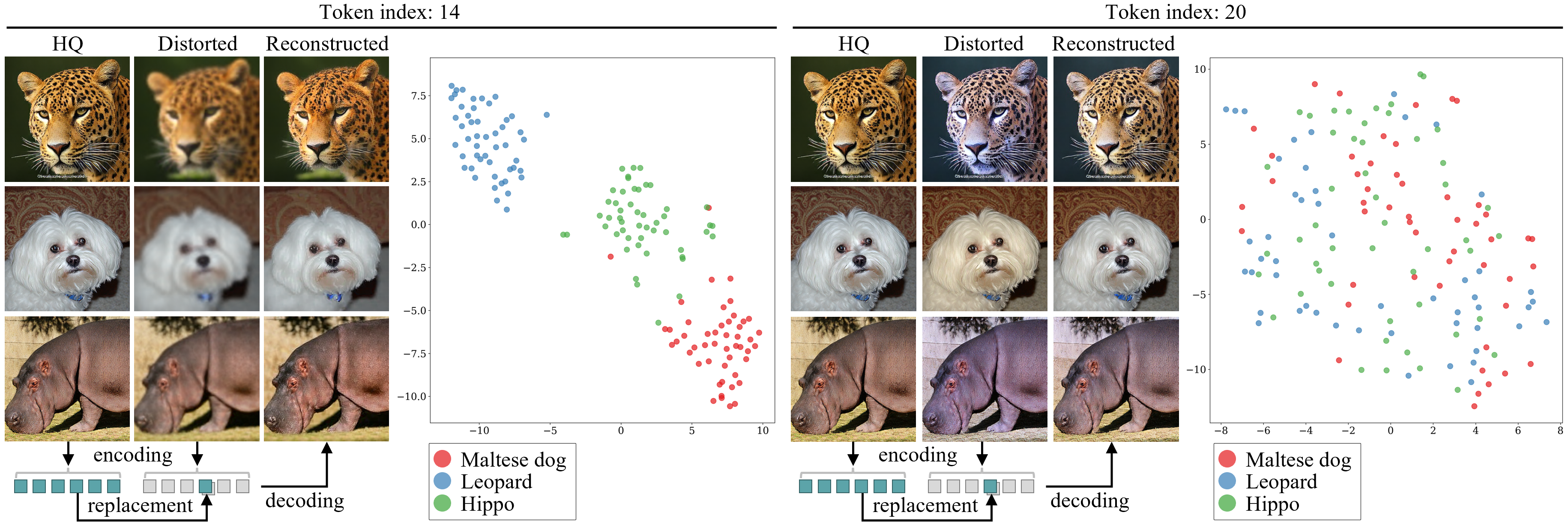}
  \caption{\textbf{Correspondence between token-wise visual effects and latent-space properties.} \textbf{Token attribution:} Reconstructed images are obtained by replacing tokens at specific indices in a distorted input with those from the corresponding HQ image using the TiTok-64 tokenizer~\cite{titok}. Token 14 primarily affects fine details (\eg, skin texture), while token 20 influences more global color and tone. \textbf{Latent space visualization:} We visualize the t-SNE distribution of latent features extracted from specific token indices of HQ images across three ImageNet classes.
  }
  \label{fig:tsne}
\end{figure}

Real-world images frequently suffer from degradation due to various factors such as transmission loss, sensor limitations, or poor shooting conditions. Under such conditions, essential high-frequency details are often lost, leading to significant performance drops in high-level vision tasks, such as image classification, object detection, and semantic segmentation~\cite{dai2016image,sr4ir,pei2018does,wu2024unsupervised}. While traditional image restoration (IR) methods~\cite{swinir, wu2024seesr, lin2024diffbir} prioritize perceptual quality, they do not necessarily guarantee machine recognizability. Consequently, task-driven image restoration (TDIR) has emerged to improve downstream model performance by preserving task-relevant semantic cues ~\cite{haris2021task,sr4ir,unirestore,edtr}. 

Recently, diffusion-based generative priors~\cite{stablediff,lin2024diffbir} have gained attention in TDIR due to their ability to robustly recover recognition-critical structures under complex degradations~\cite{unirestore,edtr}. In these pipelines, a tokenizer~\cite{vae} encodes the degraded input into a two-dimensional (2D) grid of latent tokens, and a denoising model updates all tokens to produce the restored output (\cref{fig:fig1}(a)). However, such full-token updates are suboptimal for two reasons: (i) they can unnecessarily alter reliable cues already preserved in the degraded input, causing semantic drift that harms task performance; and (ii) they increase computational cost by updating tokens with limited contribution to the downstream objective. Therefore, selective token updating is desirable in TDIR, as it allows computation to focus on task-relevant tokens while preserving reliable cues in the input.

Nevertheless, selectively updating tokens is structurally challenging in conventional 2D latent grids. In such representations, each token typically corresponds to a local image patch, where multiple factors—such as texture, boundaries, and illumination—are spatially entangled. As a result, isolating and selecting only task-relevant attributes at the token level becomes difficult, and selective updates are often reduced to coarse, spatially localized modifications.

To overcome this limitation, we turn to recently proposed one-dimensional (1D) tokenizers~\cite{titok,he2025flowtok,kim2025democratizing,video1d-1}. As shown in \cref{fig:fig1}(b), a 1D tokenizer compresses an image into a fixed-length sequence of $K$ tokens. Prior work~\cite{highlycomp} suggests that these tokens can exhibit index-wise specialization, where different indices show different sensitivities to visual attributes. 
Here, a token index refers to its sequence position. This index-structured representation provides a natural handle for selecting and refining only task-relevant tokens in TDIR. Building on TiTok~\cite{titok}, we further probe token indices and their associated visual attributes (\cref{fig:tsne}). For instance, replacing token 14 primarily affects fine details such as skin texture, whereas token 20 tends to influence more global appearance statistics such as color and tone. Consistent with these observations, the t-SNE visualization in \cref{fig:tsne} indicates that token 20 is less class-structured and appears to encode relatively generic information. In contrast, token 14 exhibits more separated clusters, suggesting stronger alignment with content-specific variations and fine-grained detail cues such as texture. Therefore, for TDIR, it is reasonable to prioritize refining tokens that encode these semantically specific, fine-detail cues, as they are more likely to directly contribute to task performance, which in turn motivates our selective token refinement strategy.

\begin{figure}[t!]
  \centering
  \includegraphics[width=\textwidth]{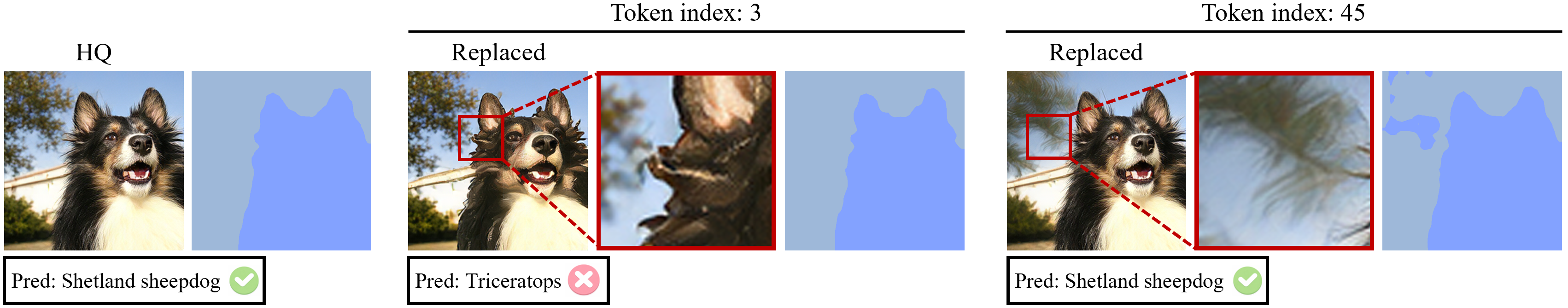}
  \caption{\textbf{Task-dependent sensitivity of latent tokens.} Token-replaced images and task predictions are obtained by replacing a single token at a specific index in HQ images with a randomly selected token. Replacing a classification-specific token degrades classification performance, whereas replacing a segmentation-specific token degrades segmentation performance. Representative visual effects include token 3 altering material appearance and token 45 modulating background blur. 
  }
  \label{fig:task_att}
\end{figure}

Furthermore, the set of task-relevant tokens varies across downstream tasks. As illustrated in \cref{fig:task_att}, perturbing tokens that primarily modulate material appearance tends to have a larger impact on classification performance, whereas perturbing tokens that modulate background blur tends to affect semantic segmentation more strongly. In other words, determining which tokens should be restored is inherently task-dependent, motivating a strategy that selectively updates task-relevant token subsets.

Based on these observations and empirical analyses, we propose TaskTok, a novel framework that selectively restores only task-relevant tokens (\cref{fig:fig1}(b)). Unlike existing large-scale generative models, TaskTok is distinguished by two key components: (i) a token switch mechanism that learns to select which tokens to restore during training, and (ii) a lightweight refinement module designed to refine the selected tokens. We evaluate TaskTok on three high-level vision tasks—image classification, object detection, and semantic segmentation—under complex degradation scenarios. As shown in~\cref{fig:fig1}(c), TaskTok achieves highly competitive task performance while restoring significantly fewer tokens than existing methods~\cite{edtr,titok}. Our main contributions are summarized as follows:

\begin{compactitem}[$\bullet$]
    \item We propose TaskTok, a TDIR framework that selectively refines task-relevant tokens via a learnable token switch and a lightweight token refinement module.
    \item We present a text-guided token attribution analysis that maps token-wise variations to natural-language attributes for interpreting task-selected tokens.
    \item We demonstrate that TaskTok significantly enhances downstream task performance with high computational efficiency.
\end{compactitem}

\section{Related Work}
\label{sec:related}
\noindent\textbf{Task-driven image restoration.}
TDIR aims to recover recognition-critical cues under degradations, rather than optimizing only pixel fidelity or perceptual quality\cite{haris2021task,liu2022image,sun2022rethinking,wu2024unsupervised}. Since attaching a generic restorer as a simple pre-processing step often yields limited downstream gains \cite{dai2016image,li2017aod,pei2018does}, recent approaches couple restoration with task networks, training them with task losses or feature-level guidance \cite{son2020urie,liu2020connecting,huang2020dsnet,li2023detection}. For instance, SR4IR~\cite{sr4ir} leverages a perceptual loss within the task network's feature space to recover task-relevant details and stabilizes training with tailored strategies. More recently, EDTR~\cite{edtr} introduces diffusion priors~\cite{stablediff} into TDIR for complex degradations. Despite this progress, existing methods still update entire latent representations (tokens or grids). This is computationally expensive and can unnecessarily overwrite reliable cues already preserved in the degraded input. In contrast, our approach explicitly estimates task-dependent token importance and selectively refines only a subset of task-relevant tokens, rather than uniformly updating all latent representations.

\noindent\textbf{1D tokenization.}
Many image generation methods~\cite{vqgan,stablediff,tian2024visual,wu2024seesr,lin2024diffbir} rely on tokenization to compress images into a low-dimensional latent space for efficient processing. Conventional image tokenizers~\cite{vae,vqvae,vqvae2} rely on 2D latent grids that preserve spatial layouts. Since this naturally ties each token to a specific local region, it becomes difficult to disentangle and selectively manipulate task-relevant attributes. In contrast, TiTok~\cite{titok} introduces a 1D tokenizer that removes the fixed correspondence between tokens and 2D patches, learning a compact latent sequence without an explicit spatial arrangement and significantly reducing the number of tokens for efficient processing. Prior work~\cite{highlycomp} further observes that TiTok-style 1D token sequences exhibit position-dependent semantics, where different token indices tend to specialize in different visual attributes. Building on this line of research, 1D tokenization has been extended to text-conditioned image generation~\cite{kim2025democratizing,he2025flowtok} and to compact video tokenization~\cite{video1d-1,tan2025sweettok,video1d-2} for efficient synthesis. Motivated by the index-wise specialization property of 1D tokens, we operate in a compact 1D token space~\cite{titok} and selectively refine only a subset of task-relevant tokens for TDIR.

\section{Proposed Method}
\label{sec:method}

\subsection{Preliminary}
TiTok~\cite{titok} is a transformer-based model that transforms an image into a compact 1D latent representation, distinct from traditional 2D grid-based tokenizers. This approach removes spatial redundancies inherent in images, enabling more efficient compression. At its core, TiTok compresses an image into a fixed number $K$ of 1D latent tokens, where the $i$-th token is denoted as $\mathbf{z}_i \in \mathbb{R}^{D}$ and $\mathbf{Z}=[\mathbf{z}_1;\dots;\mathbf{z}_K] \in \mathbb{R}^{K \times D}$. In the tokenization phase, the input image $\mathbf{I} \in \mathbb{R}^{H \times W \times 3}$ is processed by a Vision Transformer (ViT)~\cite{dosovitskiy2020image} encoder to produce the compact latent embeddings $\mathbf{Z}$. Unlike standard methods that preserve the 2D spatial grid, the encoder aggregates image information directly into these 1D tokens. 

In the de-tokenization phase, this latent representation $\mathbf{Z}$ is quantized using a codebook before being fed into the ViT decoder. The decoder then predicts the entire pixel space based on the semantic information compressed within these tokens to reconstruct the image $\hat{\mathbf{I}}$. The overall process of encoding the image into latent tokens and subsequently reconstructing it can be formally expressed as:

\begin{gather}
    \mathbf{Z} = \mathtt{Enc}(\mathbf{I}), \\
    \hat{\mathbf{I}} = \mathtt{Dec}(\mathtt{Quant}(\mathbf{Z})).
\end{gather}

Here, $\mathtt{Enc}$ and $\mathtt{Dec}$ denote the ViT encoder and decoder, respectively, and $\mathtt{Quant}$ represents the vector quantization operation that maps $\mathbf{Z}$ to the nearest entries in a learnable codebook.

\begin{figure}[t!]
  \centering
  \includegraphics[width=\textwidth]{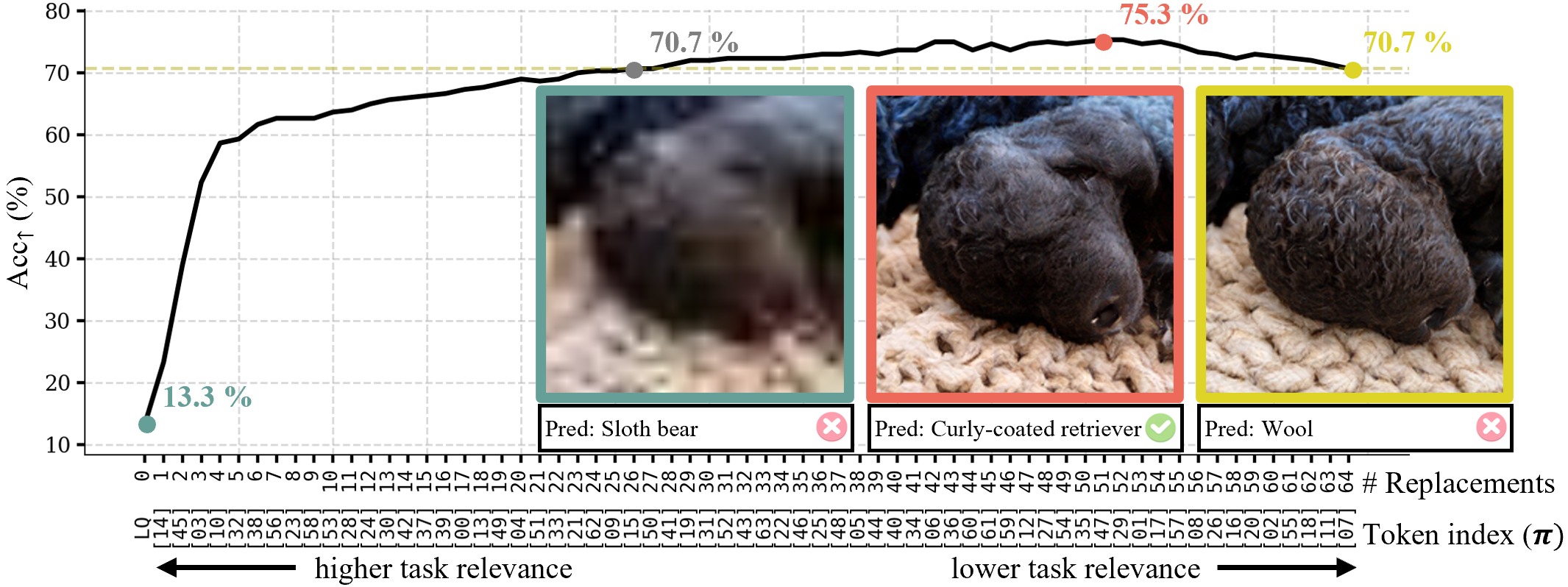}
  \caption{\textbf{Impact of token contribution to classification accuracy.} We measure
    Top-1 accuracy on the ImageNet classification task by progressively replacing LQ tokens with HQ counterparts individually. Here, tokens are encoded via TiTok-64~\cite{titok}, where LQ images are degraded by downsampling, blur,
    noise, and JPEG artifacts. The token replacement order follows the greedy-derived ranking $\boldsymbol{\pi}$ (\cref{sec:motivation}). The curve demonstrates that task performance rapidly saturates at an early stage of token replacement.
    }
  \label{fig:greedy}
\end{figure}

\subsection{Motivation: A Subset of Tokens Dominates Task Performance}
\label{sec:motivation}

To identify task-relevant tokens in TDIR, we conduct a token-level contribution analysis. We measure ImageNet~\cite{imagenet} Top-1 accuracy using TiTok-64~\cite{titok}, which encodes each image into a sequence of 64 tokens. For our analysis, we sample 300 images from the ImageNet training set and extract two sets of token sequences for each image: a high-quality (HQ) sequence, $\mathbf{Z}^{HQ}$, from the clean input, and a low-quality (LQ) sequence, $\mathbf{Z}^{LQ}$, from its degraded counterpart.

\noindent\textbf{Greedy token-importance ordering.}
To determine the relative importance of each token without exhaustive search, we employ a greedy search strategy. Starting with the all-HQ sequence, we iteratively replace each remaining HQ token with its LQ counterpart. At each step, we identify the token index whose HQ$\rightarrow$LQ replacement yields the highest resulting Top-1 accuracy—indicating the least task relevance—and permanently apply the replacement. We repeat this process until all HQ tokens are replaced by LQ tokens. Since the greedy search identifies the least task-relevant tokens first, we reverse the resulting order to obtain the final ranking $\boldsymbol{\pi}$ (from most to least task-relevant). This procedure is detailed in the supplementary material.

\noindent\textbf{Task-relevant tokens are key to performance recovery.}
We progressively restore tokens from LQ to HQ following the importance ranking $\boldsymbol{\pi}$ and track the Top-1 accuracy (\cref{fig:greedy}). The resulting curve reveals that accuracy rises sharply with the restoration of just a small subset of task-relevant tokens, after which improvements saturate. This demonstrates that classification performance is primarily driven by a few dominant tokens, while the majority contribute only marginally. Interestingly, we observe that accuracy peaks before all tokens are restored. This implies that representations optimized for faithful image reconstruction are not always optimal for downstream tasks, as certain high-frequency details may act as spurious cues. These findings strongly motivate a selective restoration strategy that prioritizes computation for task-relevant tokens over uniform refinement. We observe a similar trend for other downstream tasks, such as object detection and semantic segmentation; results are provided in the supplementary material.

\subsection{TaskTok Architecture}
\label{sec:arch}

\begin{figure}[t!]
  \centering
  \includegraphics[width=\textwidth]{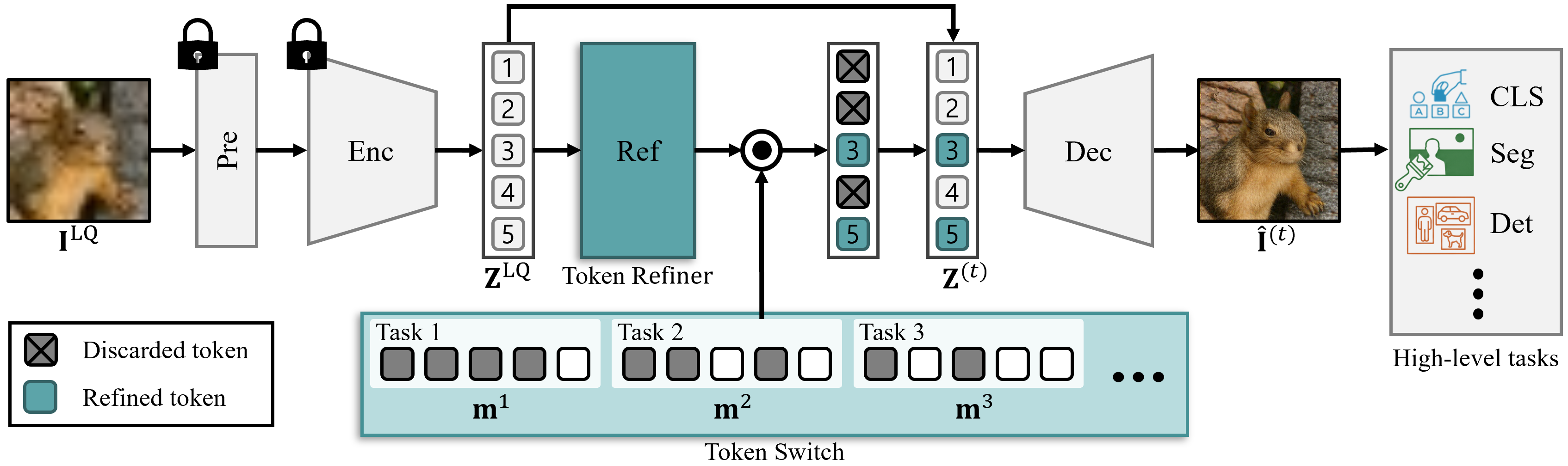}
  \caption{\textbf{Overview of TaskTok,} consisting of two key modules: a lightweight token refiner module that restores degraded representations in the 1D token space and a token switch module that determines task-specific binary masks for selective token updating. See \cref{sec:arch} for details on selective token refinement and architectural configurations.
  }
  \label{fig:method}
\end{figure}

Unlike conventional restoration methods~\cite{edtr,wu2024seesr,lin2024diffbir} that uniformly process all latent tokens, our approach selectively refines only task-relevant tokens within a 1D token space. We leverage TiTok~\cite{titok} to encode images into a fixed-length 1D latent token sequence. An overview of the proposed TaskTok framework is illustrated in~\cref{fig:method}. Since TiTok is trained on clean images, directly tokenizing inputs with complex degradations (\eg, JPEG artifacts, severe blur, noise) yields distorted latent representations. To address this, we apply a lightweight SwinIR~\cite{swinir}-based pre-restoration module $\mathtt{Pre}(\cdot)$ before tokenization and keep it frozen.

Given a degraded input $\mathbf{I}^{\mathrm{LQ}}$, the LQ token sequence is obtained as:
\begin{equation}
\mathbf{Z}^{\mathrm{LQ}}=\mathtt{Enc}\!\left(\mathtt{Pre}(\mathbf{I}^{\mathrm{LQ}})\right).
\label{eq:arch_tokenize}
\end{equation}
Here, the TiTok encoder $\mathtt{Enc}(\cdot)$ remains frozen throughout training to preserve stable, index-wise visual attributes of the pre-trained token representations.
The token refiner $\mathtt{Ref}(\cdot)$ outputs refined token representations from $\mathbf{Z}^{\mathrm{LQ}}$, while the token switch $\mathtt{Sw}(\cdot)$ determines which token is refined or preserved for every token. 

\noindent\textbf{Token refiner.} The token refiner is a lightweight module designed to restore degraded representations in the 1D token space, taking the LQ token sequence $\mathbf{Z}^{\mathrm{LQ}}$ as input. Specifically, the token refiner is implemented with a small number of transformer layers (\eg, 6), leveraging global token-to-token context to restore token-wise semantic attributions. The refined token sequence $\tilde{\mathbf{Z}}$ is given as $\tilde{\mathbf{Z}} = \mathtt{Ref}(\mathbf{Z}^{\mathrm{LQ}})$.

\noindent\textbf{Token switch.} While the greedy search (\cref{sec:motivation}) provides an importance ordering $\boldsymbol{\pi}$, relying solely on a fixed heuristic is potentially suboptimal. We therefore leverage the greedy ranking to initialize the token switch for task-relevant token selection.

The token switch optimizes a task-specific learnable probability vector $\mathbf{p}^{(t)} \in [0, 1]^K$ for each task $t\in\{1,\dots,T\}$. To obtain the binary decision mask $\mathbf{m}^{(t)}\in\{0,1\}^K$, we apply a binarization operation on $\mathbf{p}^{(t)}$ (\eg, thresholding at 0.5). Since this operation is non-differentiable, we employ the straight-through estimator during training. Given the input LQ and refined token sequences, $\mathbf{Z}^{\mathrm{LQ}}$ and $\tilde{\mathbf{Z}}$, the token switch for task $t$, denoted as $\mathtt{Sw}^{(t)}$, determines the final token sequence $\mathbf{Z}^{(t)}$ as follows:
\begin{equation}
\mathbf{Z}^{(t)}=
\mathtt{Sw}^{(t)}\!\left(
\mathbf{Z}^{\mathrm{LQ}},\ 
\tilde{\mathbf{Z}}
\right)
=\mathbf{m}^{(t)}\odot\tilde{\mathbf{Z}}+\bigl(1-\mathbf{m}^{(t)}\bigr)\odot\mathbf{Z}^{\mathrm{LQ}}.
\end{equation}

To facilitate stable training, we initialize the switch probabilities using the greedy order $\boldsymbol{\pi}^{(t)}$ for each downstream task $t$. Specifically, we linearly map the rank of each token to a probability $p_k^{(t)}$, assigning $1$ to the most important token and $0$ to the least:
\begin{equation}
p_k^{(t)}=\frac{K-r_k^{(t)}}{K-1}, \quad r_k^{(t)}\in\{1,\dots,K\},
\end{equation}
where $r_k^{(t)}$ denotes the rank of the $k$-th token in $\boldsymbol{\pi}^{(t)}$.

While hard quantization favors perceptual image quality, it introduces quantization errors that lead to semantic shift. Thus, we bypass codebook mapping and directly feed the selectively refined continuous features into the decoder. Finally, the restored image for task $t$, $\hat{\mathbf{I}}^{(t)}$, is obtained as:
\begin{equation}
\hat{\mathbf{I}}^{(t)}=\mathtt{Dec}\!\left(\mathbf{Z}^{(t)}\right).
\label{eq:switch_decode}
\end{equation}

\subsection{Training Framework}
\label{sec:training}

We adopt the two-stage training protocol of EDTR~\cite{edtr}, which decouples the optimization of the restoration network and the task network for stable training. We train TaskTok jointly across all target tasks.

\noindent \textbf{Stage 1: training TaskTok.}
We optimize TaskTok while freezing the downstream task networks. Following EDTR, we use the high-level feature distillation loss $\mathcal{L}_{HLF}$~\cite{edtr}, which minimizes the feature discrepancy between the restored output and its HQ counterpart in the task network's feature space. In addition, we introduce two losses tailored for selective token restoration. First, we apply token-level supervision only to the selected tokens using the task-specific binary mask $\mathbf{m}^{(t)}$:

\begin{equation}
\mathcal{L}_{tok}^{(t)}
=\frac{1}{\|\mathbf{m}^{(t)}\|_{1}}
\left\|\mathbf{m}^{(t)}\odot\left(\tilde{\mathbf{Z}}-\mathbf{Z}^{HQ}\right)\right\|_{1},
\qquad 
\tilde{\mathbf{Z}}=\mathtt{Ref}(\mathbf{Z}^\mathrm{LQ}).
\end{equation}
Second, to prevent the model from trivially selecting all tokens and to concentrate computation on task-relevant tokens, we minimize the average selection probability for each task $t$:
\begin{equation}
    \mathcal{L}_{mask}^{(t)}=\frac{1}{K}\sum_{k=1}^{K} p_{k}^{(t)}.
    \label{eq:loss_mask}
\end{equation}
We jointly optimize all target tasks by minimizing the expected per-task loss; in practice, we sample a task $t$ per iteration and optimize $\mathcal{L}^{(t)}_{stage1}$:

\begin{equation}
\mathcal{L}^{(t)}_{stage1}
=\mathcal{L}_{HLF}^{(t)}
+\lambda_{tok}\mathcal{L}_{tok}^{(t)}
+\lambda_{mask}\mathcal{L}_{mask}^{(t)},
\end{equation}
where $\lambda_{tok}$ and $\lambda_{mask}$ balance the loss terms.

\noindent \textbf{Stage 2: training the task network.}
We freeze TaskTok and optimize the downstream task networks. We follow EDTR and train each task network with the standard task loss $\mathcal{L}_{task}$ (\eg, cross-entropy) together with the feature matching loss $\mathcal{L}_{FM}$~\cite{edtr} to distill knowledge from the HQ-pretrained teacher network.

\section{Experiments}
We evaluate TaskTok under degraded inputs across three representative high-level vision tasks: image classification (CLS), semantic segmentation (Seg), and object detection (Det). Our goal is twofold: (i) to measure how well TaskTok recovers downstream task performance from LQ inputs, and (ii) to verify that selective, task-relevant token restoration reduces unnecessary token updates and yields a better performance–efficiency trade-off.

\noindent \textbf{Datasets.}
We use standard benchmarks for each task: ImageNet~\cite{imagenet} for classification, and PASCAL VOC2012~\cite{pascal} for semantic segmentation and object detection. To assess generalization, we additionally report classification results on unseen datasets, CUB200~\cite{cub200} and Oxford-IIIT Pet~\cite{parkhi2012cats}.

\noindent \textbf{Degradation settings.}
Following prior works~\cite{zhou2022towards,edtr}, we consider two composite degradation settings:
\begin{compactitem}[$\bullet$]
    \item Mix-A: downsampling ($s=8$) + JPEG compression ($q=75$)
    \item Mix-B: downsampling ($s\in[1,16]$) + Gaussian blur ($\sigma\in[0,8]$) + Gaussian noise ($\sigma\in[0,10]$) + JPEG compression ($q\in[50,100]$)
\end{compactitem}

\noindent \textbf{Metrics.}
We use standard task metrics: Top-1 accuracy (Acc) for classification, mIoU for segmentation, and mAP for detection. For visual quality, we report PSNR and LPIPS~\cite{lpips}. For efficiency, we report the number of restored tokens (\#Tok) and throughput, measured on an NVIDIA RTX 4090 GPU using $512\times512$ inputs.

\noindent \textbf{Downstream architectures.}
We employ representative task networks for each downstream task: We use ResNet~\cite{he2016deep} for classification, DeepLabV3 \cite{chen2017deeplab} for semantic segmentation, and Faster-RCNN~\cite{ren2015faster} for object detection. Unless stated otherwise, we additionally evaluate classification using unseen backbones (ViT~\cite{dosovitskiy2020image} and ConvNeXt~\cite{liu2022convnet}) that are not used during TaskTok training.

\begin{table*}[!t]
\captionof{table}{
\textbf{Performance across high-level vision tasks, visual quality, and efficiency under composite degradations.}
\#Tok denotes the number of restored tokens; for pixel/feature-space restorers (\eg, SwinIR, SR4IR) that update dense representations, \#Tok is not applicable. Best is in \textbf{bold} and second best is \underline{underlined}
}
\centering
\scriptsize
\setlength{\tabcolsep}{1.9pt}
\renewcommand{\arraystretch}{1.05}

\begin{tabular}{l l cl cl cl cc r}
\toprule
& \multirow{2}{*}{IR Methods}
& \multicolumn{2}{c}{CLS}
& \multicolumn{2}{c}{Seg}
& \multicolumn{2}{c}{Det}
& \multicolumn{2}{c}{Visual Quality}
& \multirow{2}{*}{\shortstack{Throughput\\(img/s)$\uparrow$}} \\
\cmidrule(lr){3-4}\cmidrule(lr){5-6}\cmidrule(lr){7-8}\cmidrule(lr){9-10}
&
& Acc$\uparrow$ & \#Tok
& mIoU$\uparrow$ & \#Tok
& mAP$\uparrow$ & \#Tok
& PSNR$\uparrow$ & LPIPS$\downarrow$
& \\
\midrule

& Oracle (Clean)
& 66.2 & ---    & 67.0 & ---   & 36.9 & ---   & ---  & ---   & ---  \\
\midrule

\multirow{8}{*}{\cellcolor{white}\rotatebox{90}{Mix-A}}
& \textit{No} restoration
& 45.9 & ---   & 42.9 & ---     & 18.0 & ---   & 22.17 & 0.6383 & ---  \\
& SwinIR~\cite{swinir}
& 50.4 & ---    & 53.8 & ---    & 22.1 & ---    &  \bf{23.28} & 0.4095 & \underline{104.9} \\
& SR4IR~\cite{sr4ir}
& 51.4 & ---    & 54.5 & ---    & 24.1 & ---    & 22.04 & 0.2163 & \underline{104.9} \\
& TiTok-64~\cite{titok}
& 52.2 & 64     & 56.8 & 64     & 23.6 & 64     & 18.02 & 0.2684 & \bf{129.3} \\
& TiTok-256~\cite{titok}
& 56.6 & 256   & 62.1 & 256    & 26.5 & 256    & 19.67 & 0.2257 & 88.7 \\
& EDTR~\cite{edtr}
& 55.7 & 4096   & 62.2 & 4096  & \underline{29.2} & 4096   &  \underline{22.20} &  \bf{0.2099} & 7.2 \\

\oursrow
& \textbf{TaskTok-64 (Ours)}
& \underline{58.1} & 12     & \underline{63.3} & 22     & 26.8  & 23    & 19.23 & 0.2427 & 75.4 \\
\oursrow
& \textbf{TaskTok-256 (Ours)}
&  \bf{59.9} & 34    &  \bf{63.9} & 79 &  \bf{30.1} & 78 & 20.53 & \underline{0.2123} & 59.9 \\
\midrule

\multirow{8}{*}{\cellcolor{white}\rotatebox{90}{Mix-B}}
& \textit{No} restoration
& 42.9 & ---    & 40.2 & ---    & 16.7   & ---   & 21.79 & 0.6354 & --- \\
& SwinIR~\cite{swinir}
& 46.4 & ---   & 49.9 & ---    & 20.1  & ---   & \bf{23.46} & 0.4997 & \underline{104.9} \\
& SR4IR~\cite{sr4ir}
& 46.8 & ---    & 51.7 & ---   & 23.1 & ---    & \underline{22.45} & 0.2482 & \underline{104.9} \\
& TiTok-64~\cite{titok}
& 47.4 & 64    & 55.5 & 64    & 22.3 & 64   & 18.12 & 0.2893 & \bf{129.3} \\
& TiTok-256~\cite{titok}
& 52.5 & 256   & 60.4 & 256   & 25.7 & 256  & 19.51 & 0.2470 & 88.7 \\
& EDTR~\cite{edtr}
& 51.6 & 4096  & 60.9 & 4096  & \underline{28.1} & 4096 & 22.28 & \bf{0.2310} & 7.2 \\

\oursrow
& \textbf{TaskTok-64 (Ours)}
& \underline{55.3} & 12    & \underline{61.9} & 22    & 25.9 &  23   & 19.12    & 0.2595 &  75.4 \\
\oursrow
& \textbf{TaskTok-256 (Ours)}
& \bf{56.2} & 34   & \bf{62.7} & 79    & \bf{28.6} & 78   & 20.45 & \underline{0.2338} & 59.9 \\
\bottomrule
\end{tabular}%
\label{table:main-table}

\centering
  \includegraphics[width=.975\textwidth]{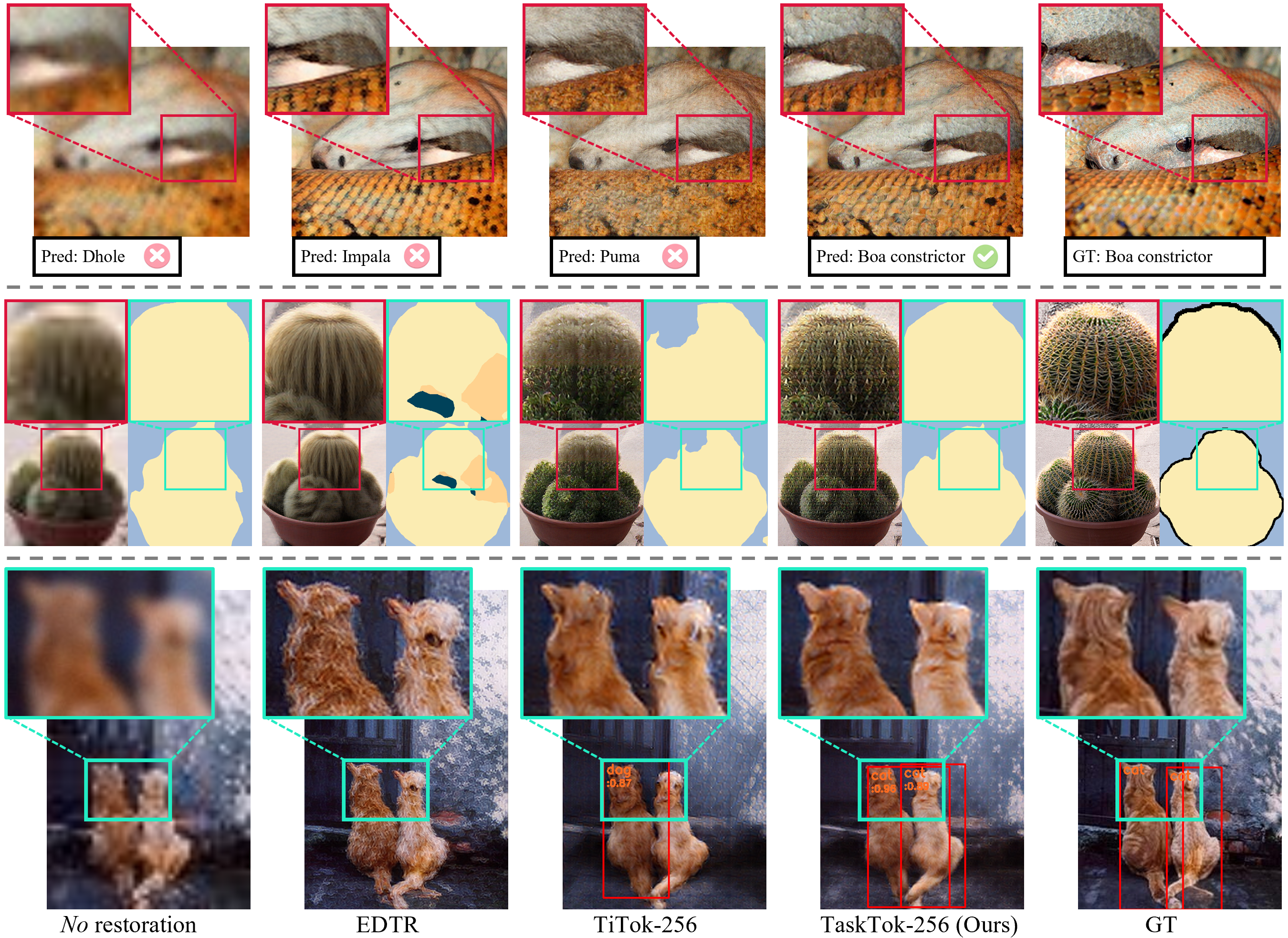}
  \captionof{figure}{\textbf{Qualitative results.}
Top/middle/bottom: image classification, semantic segmentation, and object detection.}
  \label{fig:main-fig}
\end{table*}

\subsection{Main Results}

\cref{table:main-table} and \cref{fig:main-fig} present quantitative and qualitative results under two composite degradation settings (Mix-A/Mix-B) across image classification, semantic segmentation, and object detection. Here, TiTok-$n$ and TaskTok-$n$ denote models that encode an input image into $n$ tokens, and visual quality is measured on the classification dataset (ImageNet~\cite{imagenet}). As shown in \cref{table:main-table}, the proposed TaskTok consistently outperforms prior methods across all three tasks.

\noindent \textbf{Classification.} In terms of classification, TaskTok achieves higher accuracy while restoring substantially fewer tokens than EDTR and TiTok. For instance, under the Mix-B setting, TaskTok-256 restores only $34$ tokens compared to EDTR and TiTok-256, improving Top-1 accuracy by $+4.6\%$ and $+3.7\%$, respectively. This suggests that selectively restoring and refining task-relevant tokens, rather than uniformly updating all latent representations, aligns more directly with the downstream objective. Notably, although TaskTok yields stronger task performance, it does not always achieve the best PSNR/LPIPS, indicating that optimizing conventional fidelity/perceptual metrics does not necessarily translate into better recognition. Instead, TaskTok prioritizes restoring task-relevant semantic cues while preserving reliable information already present in the input.

\noindent \textbf{Segmentation and detection.} A similar trend is observed for semantic\linebreak segmentation and object detection. Under Mix-A, for semantic segmentation, TaskTok-256 improves mIoU by $+1.7\%$ over EDTR while restoring only $79$ tokens. For object detection, TaskTok-256 improves mAP by $+0.9\%$ over EDTR with the restored token budget limited to $78$.

\noindent \textbf{Efficiency.} In terms of efficiency, TaskTok restores a very small number of tokens (\#Tok), enabling computation to focus on task-relevant representations and yielding a favorable performance--efficiency trade-off in throughput. In particular, TaskTok-256 runs $8.3\times$ faster than the one-step diffusion model, EDTR, demonstrating that selective token restoration can simultaneously improve performance and reduce computation.

\noindent \textbf{Qualitative results.} Visual comparisons in \cref{fig:main-fig} further support these findings. Without restoration, critical high-frequency details and structural cues are lost, leading to noticeable performance degradation. While EDTR and TiTok may generate visually sharper details, such details can be task-irrelevant or even misleading. In contrast, TaskTok focuses on recovering cues that matter for the downstream tasks, resulting in more consistent and reliable predictions.

\begin{table}[t]
\centering
\begin{minipage}[t]{0.485\columnwidth}
\caption{\textbf{Generalization to unseen datasets.} We evaluate Top-1 accuracy on two datasets not used during TaskTok training.}
\centering
\scriptsize
\setlength{\tabcolsep}{4pt}
\renewcommand{\arraystretch}{1.05}
\begin{tabular}{lcc}
\toprule
\multirow{2}{*}{IR Method} & \multicolumn{2}{c}{Unseen Datasets} \\
\cmidrule(lr){2-3}
 & CUB200 & Oxford-IIIT Pet \\
\midrule
\textit{No} restoration & 10.2 & 16.4 \\
TiTok-64~\cite{titok}                & 34.1 & 70.2 \\
TiTok-256~\cite{titok}                & 48.7 & 74.0 \\
EDTR~\cite{edtr}                    & \underline{50.3} & 73.7 \\
\midrule
\rowcolor{teal!10}
\textbf{TaskTok-64}    & 43.4 & \underline{75.9} \\
\rowcolor{teal!10}
\textbf{TaskTok-256}    & \bf{54.5} & \bf{79.6} \\
\bottomrule
\end{tabular}
\label{tab:unseen_data}
\end{minipage}
\hfill
\begin{minipage}[t]{0.485\columnwidth}
\caption{\textbf{Generalization to unseen classifier architectures.} We evaluate Top-1 accuracy with classification backbones not used during TaskTok training.}
\centering
\scriptsize
\setlength{\tabcolsep}{4pt}
\renewcommand{\arraystretch}{1.05}
\begin{tabular}{lcc}
\toprule
\multirow{2}{*}{IR Method} & \multicolumn{2}{c}{Unseen Backbones} \\
\cmidrule(lr){2-3}
 & ViT-B/16 & ConvNeXt-B \\
\midrule
\textit{No} restoration & 38.7  & 46.1 \\
TiTok-64~\cite{titok}                & 49.6 & 50.7 \\
TiTok-256~\cite{titok}                & \underline{55.8} & 60.3 \\
EDTR~\cite{edtr}                    & 55.3 & \underline{62.6} \\
\midrule
\rowcolor{teal!10}
\textbf{TaskTok-64}    & 55.5 & 60.9 \\
\rowcolor{teal!10}
\textbf{TaskTok-256}    & \bf{58.4} & \bf{66.5} \\
\bottomrule
\end{tabular}
\label{tab:unseen_arch}
\end{minipage}
\end{table}

\subsection{Generalizability of TaskTok}
\noindent \textbf{Cross-dataset generalization.} \cref{tab:unseen_data} reports Top-1 classification accuracy on unseen datasets, CUB200 and Oxford-IIIT Pet, to evaluate dataset-level generalization. TaskTok demonstrates consistent robustness on unseen data. Notably, TaskTok-256 achieves the best performance on CUB200 (54.5\%), outperforming EDTR (50.3\%) by +4.2\% and TiTok-256 (48.7\%) by +5.8\%. A similar trend is observed on Oxford-IIIT Pet, indicating that TaskTok effectively generalizes to diverse data distributions without overfitting.

\noindent \textbf{Cross-architecture generalization.} \cref{tab:unseen_arch} further evaluates whether the gains persist on unseen classifier architectures that were not used during TaskTok training. We measure Top-1 accuracy using classification backbones pretrained on HQ images (\eg, ViT-B/16 and ConvNeXt-B). TaskTok maintains stable improvements across these unseen models; for instance, TaskTok-256 achieves 66.5\% Top-1 accuracy with ConvNeXt-B. These findings confirm that TaskTok learns robust, generalizable restoration representations independent of specific classifier designs. All generalization results are reported under Mix-B.

\begin{table*}[t]
\caption{\textbf{Ablation study on TaskTok components.}
Pre: pre-restoration module; Ref: token refiner; Sw: token switch;
$\boldsymbol{\pi}$: importance order initialization; VQ: codebook quantization.}
\centering
\scriptsize
\setlength{\tabcolsep}{2.5pt}
\renewcommand{\arraystretch}{1.15}
\begin{tabular}{lccccccccccccc}
\toprule
\multirow{2}{*}{Variant}
& \multicolumn{7}{c}{Ablation toggles}
& \multicolumn{2}{c}{CLS}
& \multicolumn{2}{c}{Seg}
& \multicolumn{2}{c}{Det} \\
\cmidrule(lr){2-8}\cmidrule(lr){9-10}\cmidrule(lr){11-12}\cmidrule(lr){13-14}
& Pre & Ref & $\mathcal{L}_{tok}$ & Sw & $\mathcal{L}_{mask}$ & $\boldsymbol{\pi}$ & VQ
& Acc$\uparrow$ & \#Tok
& mIoU$\uparrow$ & \#Tok
& mAP$\uparrow$ & \#Tok \\
\midrule
TiTok-64~\cite{titok}
& \xmark & \xmark & \xmark & \xmark & \xmark & \xmark & \cmark
& 47.4 & 64
& 55.5 & 64
& 22.3 & 64 \\
\midrule
$\boldsymbol{+}$ Pre-restoration
& \cmark & \xmark & \xmark & \xmark & \xmark & \xmark & \cmark
& 48.5 & 64
& 56.9 & 64
& 23.1 & 64 \\
$\boldsymbol{+}$ Token refiner
& \cmark & \cmark & \xmark & \xmark & \xmark & \xmark & \cmark
& 46.3 & 64
& 56.0 & 64
& 21.8 & 64 \\
$\boldsymbol{+}$ $\mathcal{L}_{tok}$
& \cmark & \cmark & \cmark & \xmark & \xmark & \xmark & \cmark
& 46.7 & 64
& 56.2 & 64
& 22.2 & 64 \\
$\boldsymbol{+}$ Token switch
& \cmark & \cmark & \cmark & \cmark & \xmark & \xmark & \cmark
& 46.9 & 31
& 57.1 & 33
& 22.1 & 35 \\
$\boldsymbol{+}$ $\mathcal{L}_{mask}$
& \cmark & \cmark & \cmark & \cmark & \cmark & \xmark & \cmark
& 50.9 & 14
& 58.3 & 26
& 22.7 & 25 \\
$\boldsymbol{+}$ $\boldsymbol{\pi}$ initialization
& \cmark & \cmark & \cmark & \cmark & \cmark & \cmark & \cmark
& \underline{53.0} & 10
& \underline{60.4} & 24
& \underline{24.3} & 22 \\
\rowcolor{teal!10}
\textbf{TaskTok-64}
& \cmark & \cmark & \cmark & \cmark & \cmark & \cmark & \xmark
& \bf{55.3} & 12
& \bf{61.9} & 22
& \bf{25.9} & 23 \\
\bottomrule
\end{tabular}
\label{tab:ablation}
\end{table*}

\begin{figure}[t!]
  \centering
  \includegraphics[width=\textwidth]{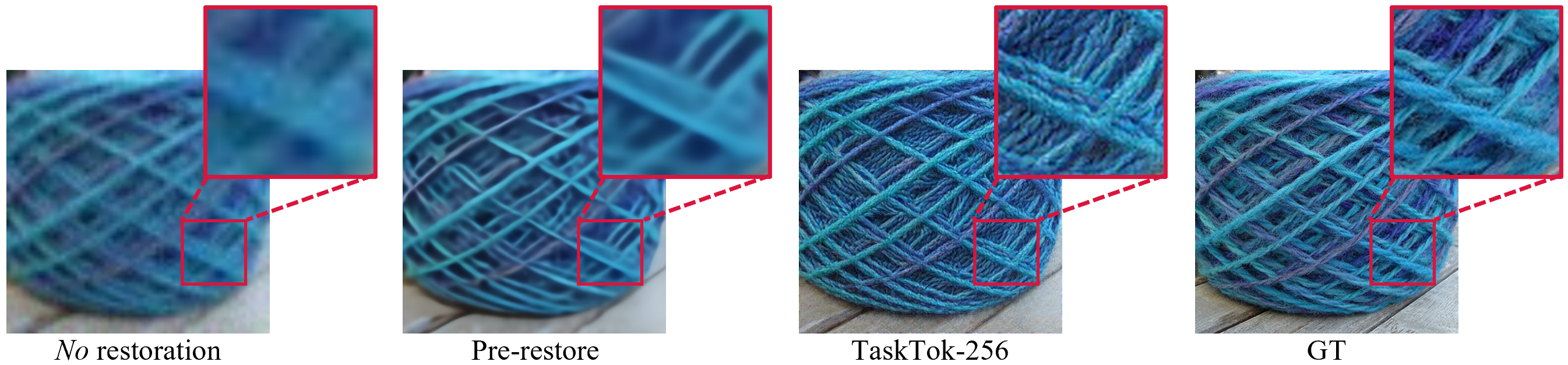}
  \caption{Qualitative examples of the pre-restoration outputs and the TaskTok results.}
  \label{fig:pre_restore}
\end{figure}

\subsection{Ablation Studies}

\cref{tab:ablation} summarizes the ablation results in the Mix-B setting, analyzing the contribution of each component step by step. We first apply a pre-restoration module before TiTok to suppress degradation artifacts. This mitigates the risk that artifacts act as misleading generation cues, and consequently improves performance across all three tasks. \cref{fig:pre_restore} presents qualitative examples of the pre-restoration outputs and the final TaskTok results. While pre-restoration mainly suppresses degradation artifacts before tokenization, TaskTok restores task-relevant cues through selective token refinement. In contrast, adding the token refiner to restore all tokens degrades performance, and incorporating $\mathcal{L}_{tok}$ yields only marginal gains. This suggests that (i) uniformly restoring every token is not always beneficial for downstream tasks, and (ii) the lightweight refiner is insufficient for fully restoring the entire token sequence. Introducing the token switch leads to a clear improvement by enabling selective updates of task-relevant tokens. In particular, with the initialization of the token order $\boldsymbol{\pi}$ and the mask regularizer $\mathcal{L}_{mask}$ to discourage selecting unnecessary tokens, the model learns an optimal token subset and achieves gains of $5.6\%$, $4.9\%$, and $2.0\%$ on CLS/Seg/Det over TiTok-64. Finally, removing quantization alleviates the semantic shift induced by quantization, yielding the best performance of the final TaskTok model.

\subsection{Overhead of the Token Refiner}
TaskTok employs a selective update strategy, refining only task-relevant tokens instead of uniformly processing all latent representations. Since its objective is to focus on refining the small subset selected by the token switch, the token refiner can be designed as a highly lightweight transformer. Consequently, for TaskTok-64 (total latency: 13.26\,ms/img), the refiner introduces a negligible overhead of just 0.08\,ms/img (0.60\%) and contributes only 19.0\,M parameters (4.6\%) to the 413.6\,M total. These results confirm that selective restoration incurs minimal cost while significantly enhancing downstream task performance.

\subsection{Cross-Task Transferability}
\begin{figure}[t]
\centering
\begin{minipage}[t]{0.425\linewidth}
  \centering
  {\captionsetup{type=table, belowskip=10pt}%
   \captionof{table}{\textbf{Cross-task token transfer.}
   Using tokens selected by a task (CLS/Seg/Det) for other tasks.}
   \label{tab:transfer}}
  \scriptsize
  \setlength{\tabcolsep}{1.1pt}
  \renewcommand{\arraystretch}{1.3}
    \begin{tabular}{lccc}
    \toprule
    \multirow{2}{*}{Token Set}
    & CLS & Seg & Det \\
    & (Acc$\uparrow$) & (mIoU$\uparrow$) & (mAP$\uparrow$) \\
    \midrule
    CLS-selected &\cellcolor{teal!10}{\bf{55.3}} & 59.9 & 24.2 \\
    Seg-selected & \underline{53.9} & \cellcolor{teal!10}{\bf{61.9}} & \underline{25.3} \\
    Det-selected & 53.5 & \underline{61.3} & \cellcolor{teal!10}{\bf{25.9}} \\
    \bottomrule
    \end{tabular}
\end{minipage}\hfill
\begin{minipage}[t]{0.565\linewidth}
  \centering
  \vspace{.5\baselineskip}
  \includegraphics[width=\linewidth]{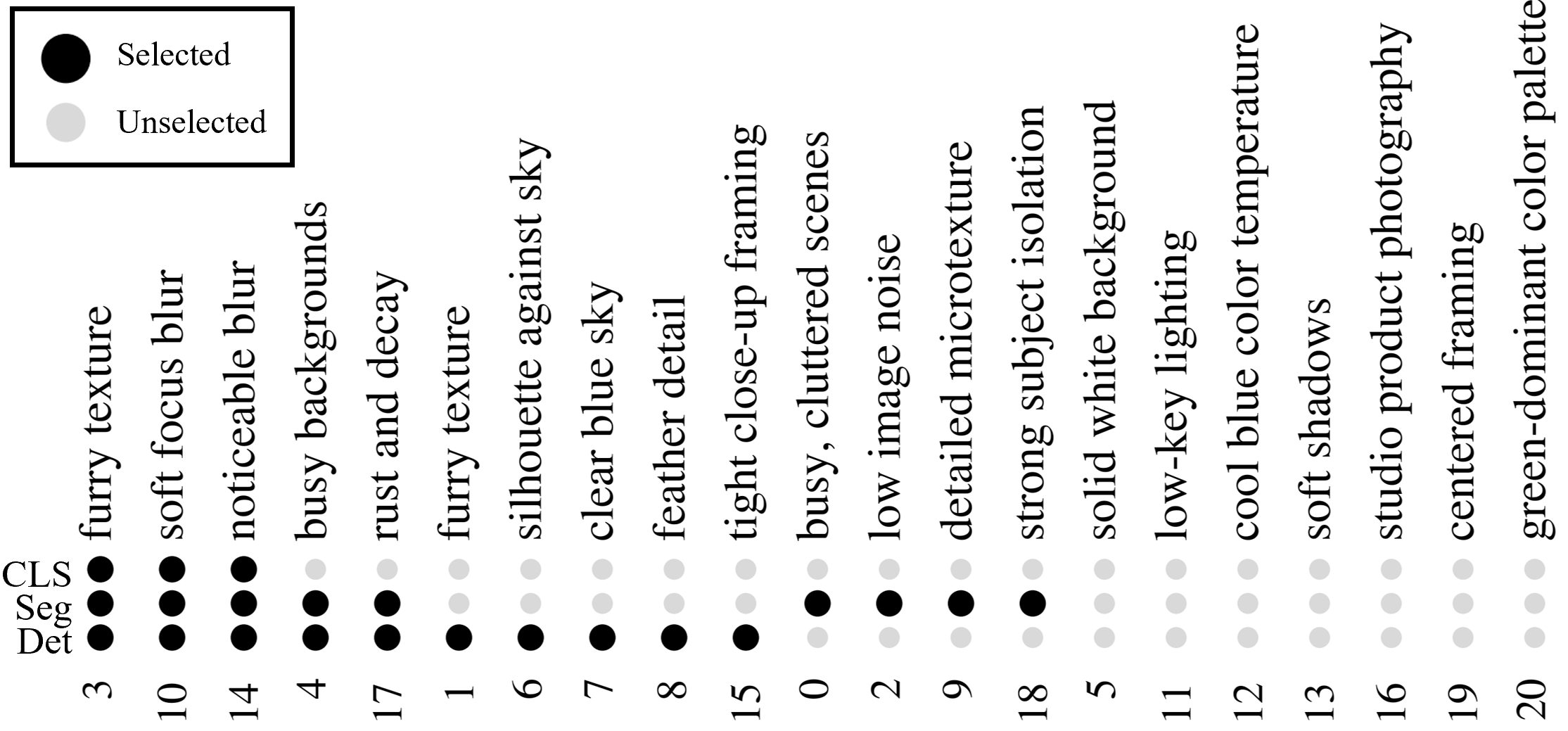}
  \caption{\textbf{Task-wise token attribute selection (TaskTok-64).} First 21 latent tokens (0--20).}
  \label{fig:token_diff20}
\end{minipage}
\end{figure}

\cref{tab:transfer}  compares the performance when a token subset selected for one task is directly applied to another. Using task-specific tokens yields the best performance for each respective task. For instance, evaluating CLS with tokens selected by Seg/Det drops the accuracy by 1.4\% / 1.8\% from 55.3\%. This clearly indicates that performance relies on whether the restored tokens are truly task-critical, rather than merely the total number of restored tokens. Interestingly, the Seg $\leftrightarrow$ Det transfer exhibits much smaller performance drops (0.6\% / 0.6\%), implying that segmentation and detection share underlying structural cues. Overall, these results demonstrate that our model successfully identifies optimal, task-specific subsets, which is crucial for maximizing downstream performance. All results are reported under Mix-B.

\begin{figure}[t!]
  \centering
  \includegraphics[width=\textwidth]{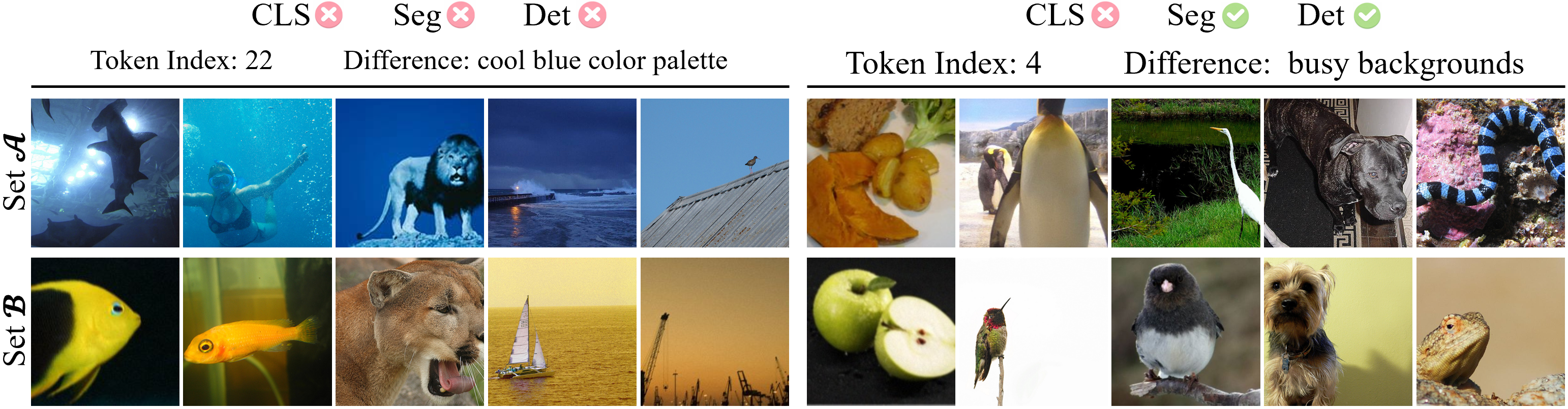}
  \caption{\textbf{Qualitative examples of text-based attribute analysis.} Images from two distinct clusters (top and bottom rows) for specific token indices. Difference indicates the core visual variation between the two sets, summarized by an image-set difference captioning method~\cite{visdiff}.
  }
  \label{fig:difference}
\end{figure}

\begin{algorithm}[!ht]
\caption{Text-Guided Token Attribution Analysis}
\label{alg:token_attr}
\KwIn{Image set $\mathcal{D}$, frozen tokenizer encoder $\mathtt{Enc}(\cdot)$, token index $k$, clustering algorithm $\mathtt{Cluster}(\cdot)$, image-set difference captioner $\mathtt{DiffCap}(\cdot,\cdot)$}
\KwOut{Proxy attribute label $c_k$ for token index $k$}

Initialize token set $\mathcal{Z}_k \leftarrow [\ ]$\;
\ForEach{$\mathbf{I}\in\mathcal{D}$}{
    Extract $\mathbf{z}_k \leftarrow \mathtt{Enc}(\mathbf{I})[k]$\;
    Append $\mathbf{z}_k$ to $\mathcal{Z}_k$\;
}
Obtain clusters $\{\mathcal{C}_1,\dots,\mathcal{C}_M\} \leftarrow \mathtt{Cluster}(\mathcal{Z}_k)$\;
Compute the center $\boldsymbol{\mu}_m$ for each cluster $\mathcal{C}_m$\;
Select $(a,b) \leftarrow \arg\max_{i \neq j} d(\boldsymbol{\mu}_i, \boldsymbol{\mu}_j), \quad 1 \le i, j \le M$\;
Define image sets $\mathcal{A}, \mathcal{B}$ from the images assigned to $\mathcal{C}_a$ and $\mathcal{C}_b$\;
Generate caption $c_k \leftarrow \mathtt{DiffCap}(\mathcal{A},\mathcal{B})$\;
\Return $c_k$\;
\end{algorithm}

\subsection{Task-Specific Token Attribution Analysis}
The visual semantics encoded by latent tokens are highly entangled in a high-dimensional space, making them difficult to define individually. To interpret TaskTok's selective update mechanism from a human-understandable semantic perspective, we conduct a text-guided attribution analysis by mapping the axes of visual variation into natural language descriptions.

\noindent \textbf{Text-based attribute analysis.} We summarize the attribution pipeline in \cref{alg:token_attr}. We use 30K clean and degraded images from the ImageNet~\cite{imagenet} validation set. For each token index, we collect its representations across images and cluster them using HDBSCAN~\cite{hdbscan}. A visual attribute can appear as multiple coherent clusters; for example, color may split into several hue-specific clusters. We therefore select the two most distant clusters and define their corresponding image sets as $\mathcal{A}$ and $\mathcal{B}$. We then follow VisDiff~\cite{visdiff} to describe their dominant visual difference: LLaVA-1.5~\cite{llava} first generates captions for both sets, GPT-5~\cite{gpt5} proposes candidate difference descriptions, and CLIP~\cite{clip} ranks them. \cref{fig:difference} provides qualitative examples demonstrating that the clustering results for each token index capture a dominant axis of visual variation, which the generated difference captions consistently summarize. This supports using the difference captions as proxy labels for the dominant visual attribute modulated by each token index.

\noindent \textbf{Task-specific token selection trends.} \cref{fig:token_diff20} summarizes the attribute analysis for the first 21 TaskTok-64 token indices alongside their selection indicators across three downstream tasks.

Our analysis shows that universally selected tokens are primarily associated with fine-grained details and morphological sharpness (\eg, furry texture, soft-focus blur, and noticeable blur), whereas consistently excluded tokens tend to reflect global appearance factors such as color (\eg, cool-blue color temperature, green-dominant color palette) or lighting. This suggests that such global attributes are either already sufficiently preserved in the degraded inputs or do not decisively impact the final performance of these high-level tasks.

Meanwhile, both Seg and Det tend to select tokens related to background clutter and scene complexity, consistent with their need for object localization and boundary delineation. Notably, requiring dense pixel-level precision, Seg exhibits a heightened sensitivity to localized detail variations, such as image noise and microtexture. Overall, this analysis offers an interpretable perspective on the visual attributes prioritized by TaskTok's selective refinement mechanism. The complete token attribution results, along with additional qualitative examples of the image-set difference analysis, are provided in the supplementary material.

\section{Conclusion}
This paper proposes TaskTok, a novel TDIR framework that selectively refines task-relevant representations in a 1D latent token space. We observe that task-specific tokens play a critical role in downstream recognition, and that indiscriminate restoration may distort semantic cues essential for machine perception. To address this, we propose a set of task-aware components, including a learnable token switch and a lightweight token refiner, to selectively restore only the token subsets that are most relevant to each downstream objective. Experimental results demonstrate that TaskTok significantly improves the performance of various high-level vision tasks, such as image classification, semantic segmentation, and object detection, under complex image degradation scenarios. We anticipate that our approach will provide new insights into task-aware and computationally efficient token processing for TDIR.

\section*{Acknowledgements}
This work was supported by the National Research Foundation of Korea (NRF) grant funded by the Korea government (MSIT) (No. RS-2025-16068196).

%
%
\bibliographystyle{splncs04}
\bibliography{main}

@String(PAMI  = {IEEE Trans. Pattern Anal. Mach. Intell.})

@String(IJCV  = {Int. J. Comput. Vis.})

@String(CVPR  = {IEEE Conf. Comput. Vis. Pattern Recog.})

@String(ICCV  = {Int. Conf. Comput. Vis.})

@String(ECCV  = {Eur. Conf. Comput. Vis.})

@String(NeurIPS = {Adv. Neural Inform. Process. Syst.})

@String(ICML  = {Int. Conf. Mach. Learn.})

@String(ICLR  = {Int. Conf. Learn. Represent.})

@String(AAAI  = {AAAI})

@String(TIP   = {IEEE Trans. Image Process.})

@String(PAMI  = {IEEE TPAMI})

@String(IJCV  = {IJCV})

@String(CVPR  = {CVPR})

@String(ICCV  = {ICCV})

@String(ECCV  = {ECCV})

@String(NeurIPS = {NeurIPS})

@String(ICML  = {ICML})

@String(ICLR  = {ICLR})

@String(TIP   = {IEEE TIP})

@inproceedings{titok,
  title={An image is worth 32 tokens for reconstruction and generation},
  author={Yu, Qihang and Weber, Mark and Deng, Xueqing and Shen, Xiaohui and Cremers, Daniel and Chen, Liang-Chieh},
  booktitle = NeurIPS,
  pages={128940--128966},
  year={2024}
}

@inproceedings{sun2022rethinking,
  title={Rethinking image restoration for object detection},
  author={Sun, Shangquan and Ren, Wenqi and Wang, Tao and Cao, Xiaochun},
  booktitle=NeurIPS,
  pages={4461--4474},
  year={2022}
}

@inproceedings{vqvae,
  title={Neural discrete representation learning},
  author={Van Den Oord, Aaron and Vinyals, Oriol and others},
  booktitle=NeurIPS,
  year={2017}
}

@inproceedings{vqvae2,
  title={Generating diverse high-fidelity images with {VQ-VAE-2}},
  author={Razavi, Ali and Van den Oord, Aaron and Vinyals, Oriol},
  booktitle=NeurIPS,
  year={2019}
}

@inproceedings{tian2024visual,
  title={Visual autoregressive modeling: Scalable image generation via next-scale prediction},
  author={Tian, Keyu and Jiang, Yi and Yuan, Zehuan and Peng, Bingyue and Wang, Liwei},
  booktitle=NeurIPS,
  pages={84839--84865},
  year={2024}
}

@inproceedings{zhou2022towards,
  title={Towards robust blind face restoration with codebook lookup transformer},
  author={Zhou, Shangchen and Chan, Kelvin and Li, Chongyi and Loy, Chen Change},
  booktitle=NeurIPS,
  pages={30599--30611},
  year={2022}
}

@inproceedings{ren2015faster,
  title={{Faster R-CNN}: Towards real-time object detection with region proposal networks},
  author={Ren, Shaoqing and He, Kaiming and Girshick, Ross and Sun, Jian},
  booktitle=NeurIPS,
  year={2015}
}

@inproceedings{llava,
  title={Visual instruction tuning},
  author={Liu, Haotian and Li, Chunyuan and Wu, Qingyang and Lee, Yong Jae},
  booktitle=NeurIPS,
  year={2023}
}

@inproceedings{dai2016image,
  title={Is image super-resolution helpful for other vision tasks?},
  author={Dai, Dengxin and Wang, Yujian and Chen, Yuhua and Van Gool, Luc},
  booktitle={WACV},
  pages={1--9},
  year={2016}
}

@inproceedings{sr4ir,
  title={Beyond image super-resolution for image recognition with task-driven perceptual loss},
  author={Kim, Jaeha and Oh, Junghun and Lee, Kyoung Mu},
  booktitle=CVPR,
  pages={2651--2661},
  year={2024}
}

@inproceedings{wu2024seesr,
  title={{SeeSR}: Towards semantics-aware real-world image super-resolution},
  author={Wu, Rongyuan and Yang, Tao and Sun, Lingchen and Zhang, Zhengqiang and Li, Shuai and Zhang, Lei},
  booktitle=CVPR,
  pages={25456--25467},
  year={2024}
}

@inproceedings{unirestore,
  title={{UniRestore}: Unified perceptual and task-oriented image restoration model using diffusion prior},
  author={Chen, I and Chen, Wei-Ting and Liu, Yu-Wei and Chiang, Yuan-Chun and Kuo, Sy-Yen and Yang, Ming-Hsuan and others},
  booktitle=CVPR,
  pages={17969--17979},
  year={2025}
}

@inproceedings{stablediff,
  title={High-resolution image synthesis with latent diffusion models},
  author={Rombach, Robin and Blattmann, Andreas and Lorenz, Dominik and Esser, Patrick and Ommer, Bj{\"o}rn},
  booktitle=CVPR,
  pages={10684--10695},
  year={2022}
}

@inproceedings{vqgan,
  title={Taming transformers for high-resolution image synthesis},
  author={Esser, Patrick and Rombach, Robin and Ommer, Bjorn},
  booktitle=CVPR,
  pages={12873--12883},
  year={2021}
}

@inproceedings{imagenet,
  title={{ImageNet}: A large-scale hierarchical image database},
  author={Deng, Jia and Dong, Wei and Socher, Richard and Li, Li-Jia and Li, Kai and Fei-Fei, Li},
  booktitle=CVPR,
  pages={248--255},
  year={2009}
}

@inproceedings{parkhi2012cats,
  title={Cats and dogs},
  author={Parkhi, Omkar M and Vedaldi, Andrea and Zisserman, Andrew and Jawahar, CV},
  booktitle=CVPR,
  pages={3498--3505},
  year={2012}
}

@inproceedings{lpips,
  title={The unreasonable effectiveness of deep features as a perceptual metric},
  author={Zhang, Richard and Isola, Phillip and Efros, Alexei A and Shechtman, Eli and Wang, Oliver},
  booktitle=CVPR,
  pages={586--595},
  year={2018}
}

@inproceedings{he2016deep,
  title={Deep residual learning for image recognition},
  author={He, Kaiming and Zhang, Xiangyu and Ren, Shaoqing and Sun, Jian},
  booktitle=CVPR,
  pages={770--778},
  year={2016}
}

@inproceedings{liu2022convnet,
  title={A convnet for the 2020s},
  author={Liu, Zhuang and Mao, Hanzi and Wu, Chao-Yuan and Feichtenhofer, Christoph and Darrell, Trevor and Xie, Saining},
  booktitle=CVPR,
  pages={11976--11986},
  year={2022}
}

@inproceedings{visdiff,
  title={Describing differences in image sets with natural language},
  author={Dunlap, Lisa and Zhang, Yuhui and Wang, Xiaohan and Zhong, Ruiqi and Darrell, Trevor and Steinhardt, Jacob and Gonzalez, Joseph E and Yeung-Levy, Serena},
  booktitle=CVPR,
  pages={24199--24208},
  year={2024}
}

@inproceedings{pei2018does,
  title={Does haze removal help {CNN}-based image classification?},
  author={Pei, Yanting and Huang, Yaping and Zou, Qi and Lu, Yuhang and Wang, Song},
  booktitle=ECCV,
  pages={682--697},
  year={2018}
}

@inproceedings{wu2024unsupervised,
  title={Unsupervised variational translator for bridging image restoration and high-level vision tasks},
  author={Wu, Jiawei and Jin, Zhi},
  booktitle=ECCV,
  pages={214--231},
  year={2024}
}

@inproceedings{lin2024diffbir,
  title={{DiffBIR}: Toward blind image restoration with generative diffusion prior},
  author={Lin, Xinqi and He, Jingwen and Chen, Ziyan and Lyu, Zhaoyang and Dai, Bo and Yu, Fanghua and Qiao, Yu and Ouyang, Wanli and Dong, Chao},
  booktitle=ECCV,
  pages={430--448},
  year={2024}
}

@inproceedings{son2020urie,
  title={{URIE}: Universal image enhancement for visual recognition in the wild},
  author={Son, Taeyoung and Kang, Juwon and Kim, Namyup and Cho, Sunghyun and Kwak, Suha},
  booktitle=ECCV,
  pages={749--765},
  year={2020}
}

@inproceedings{swinir,
  title={{SwinIR}: Image restoration using swin transformer},
  author={Liang, Jingyun and Cao, Jiezhang and Sun, Guolei and Zhang, Kai and Van Gool, Luc and Timofte, Radu},
  booktitle=ICCV,
  pages={1833--1844},
  year={2021}
}

@inproceedings{edtr,
  title={Exploiting diffusion prior for task-driven image restoration},
  author={Kim, Jaeha and Oh, Junghun and Lee, Kyoung Mu},
  booktitle=ICCV,
  pages={10151--10161},
  year={2025}
}

@inproceedings{he2025flowtok,
  title={{FlowTok}: Flowing seamlessly across text and image tokens},
  author={He, Ju and Yu, Qihang and Liu, Qihao and Chen, Liang-Chieh},
  booktitle=ICCV,
  pages={16629--16640},
  year={2025}
}

@inproceedings{kim2025democratizing,
  title={Democratizing text-to-image masked generative models with compact text-aware one-dimensional tokens},
  author={Kim, Dongwon and He, Ju and Yu, Qihang and Yang, Chenglin and Shen, Xiaohui and Kwak, Suha and Chen, Liang-Chieh},
  booktitle=ICCV,
  pages={18442--18452},
  year={2025}
}

@inproceedings{video1d-1,
  title={The best of both worlds: Integrating language models and diffusion models for video generation},
  author={Yin, Aoxiong and Tan, Xu and Shen, Kai and Leng, Yichong and Zhou, Xinyu and Li, Juncheng and Tang, Siliang},
  booktitle=ICCV,
  pages={15604--15615},
  year={2025}
}

@inproceedings{li2017aod,
  title={{AOD-Net}: All-in-one dehazing network},
  author={Li, Boyi and Peng, Xiulian and Wang, Zhangyang and Xu, Jizheng and Feng, Dan},
  booktitle=ICCV,
  pages={4770--4778},
  year={2017}
}

@inproceedings{tan2025sweettok,
  title={{SweetTok}: Semantic-aware spatial-temporal tokenizer for compact video discretization},
  author={Tan, Zhentao and Xue, Ben and Jia, Jian and Wang, Junhao and Ye, Wencai and Shi, Shaoyun and Sun, Mingjie and Wu, Wenjin and Chen, Quan and Jiang, Peng},
  booktitle=ICCV,
  pages={23541--23550},
  year={2025}
}

@inproceedings{highlycomp,
    title={Highly Compressed Tokenizer Can Generate Without Training},
    author={Lukas Lao Beyer and Tianhong Li and Xinlei Chen and Sertac Karaman and Kaiming He},
    booktitle=ICML,
    year={2025}
}

@inproceedings{clip,
  title={Learning Transferable Visual Models From Natural Language Supervision},
  author={Alec Radford and Jong Wook Kim and Chris Hallacy and A. Ramesh and Gabriel Goh and Sandhini Agarwal and Girish Sastry and Amanda Askell and Pamela Mishkin and Jack Clark and Gretchen Krueger and Ilya Sutskever},
  booktitle={ICML},
  year={2021}
}

@inproceedings{dosovitskiy2020image,
  title={An image is worth 16x16 words: Transformers for image recognition at scale},
  author={Dosovitskiy, Alexey and Beyer, Lucas and Kolesnikov, Alexander and Weissenborn, Dirk and Zhai, Xiaohua and Unterthiner, Thomas and Dehghani, Mostafa and Minderer, Matthias and Heigold, Georg and Gelly, Sylvain and others},
  booktitle=ICLR,
  year={2020}
}

@inproceedings{vae,
  title={Auto-encoding variational bayes},
  author={Kingma, Diederik P and Welling, Max},
  booktitle=ICLR,
  year={2014}
}

@inproceedings{liu2022image,
  title={Image-adaptive {YOLO} for object detection in adverse weather conditions},
  author={Liu, Wenyu and Ren, Gaofeng and Yu, Runsheng and Guo, Shi and Zhu, Jianke and Zhang, Lei},
  booktitle=AAAI,
  pages={1792--1800},
  year={2022}
}

@article{huang2020dsnet,
  title={{DSNet}: Joint semantic learning for object detection in inclement weather conditions},
  author={Huang, Shih-Chia and Le, Trung-Hieu and Jaw, Da-Wei},
  journal=PAMI,
  volume={43},
  number={8},
  pages={2623--2633},
  year={2020},
  publisher={IEEE}
}

@article{chen2017deeplab,
  title={{DeepLab}: Semantic image segmentation with deep convolutional nets, atrous convolution, and fully connected crfs},
  author={Chen, Liang-Chieh and Papandreou, George and Kokkinos, Iasonas and Murphy, Kevin and Yuille, Alan L},
  journal=PAMI,
  volume={40},
  number={4},
  pages={834--848},
  year={2017},
  publisher={IEEE}
}

@article{li2023detection,
  title={Detection-friendly dehazing: Object detection in real-world hazy scenes},
  author={Li, Chengyang and Zhou, Heng and Liu, Yang and Yang, Caidong and Xie, Yongqiang and Li, Zhongbo and Zhu, Liping},
  journal=PAMI,
  volume={45},
  number={7},
  pages={8284--8295},
  year={2023},
  publisher={IEEE}
}

@article{liu2020connecting,
  title={Connecting image denoising and high-level vision tasks via deep learning},
  author={Liu, Ding and Wen, Bihan and Jiao, Jianbo and Liu, Xianming and Wang, Zhangyang and Huang, Thomas S},
  journal=TIP,
  volume={29},
  pages={3695--3706},
  year={2020},
  publisher={IEEE}
}

@article{pascal,
  title={The {Pascal} visual object classes {(VOC)} challenge},
  author={Everingham, Mark and Van Gool, Luc and Williams, Christopher KI and Winn, John and Zisserman, Andrew},
  journal=IJCV,
  volume={88},
  pages={303--338},
  year={2010},
  publisher={Springer}
}

@article{video1d-2,
  title={Learning adaptive and temporally causal video tokenization in a {1D} latent space},
  author={Li, Yan and Tian, Changyao and Xia, Renqiu and Liao, Ning and Guo, Weiwei and Yan, Junchi and Li, Hongsheng and Dai, Jifeng and Li, Hao and Yang, Xue},
  journal={arXiv preprint arXiv:2505.17011},
  year={2025}
}

@article{gpt5,
  title={Openai {GPT}-5 system card},
  author={Singh, Aaditya and Fry, Adam and Perelman, Adam and Tart, Adam and Ganesh, Adi and El-Kishky, Ahmed and McLaughlin, Aidan and Low, Aiden and Ostrow, AJ and Ananthram, Akhila and others},
  journal={arXiv preprint arXiv:2601.03267},
  year={2025}
}

@inproceedings{haris2021task,
  title={Task-driven super resolution: Object detection in low-resolution images},
  author={Haris, Muhammad and Shakhnarovich, Greg and Ukita, Norimichi},
  booktitle={International Conference on Neural Information Processing},
  pages={387--395},
  year={2021}
}

@article{cub200,
  title={Caltech-UCSD birds 200},
  author={Welinder, Peter and Branson, Steve and Mita, Takeshi and Wah, Catherine and Schroff, Florian and Belongie, Serge and Perona, Pietro},
  journal={Technical Report California Institute of Technology. CNS-TR-2010-001.},
  year={2010}
}

@inproceedings{hdbscan,
  title={Density-based clustering based on hierarchical density estimates},
  author={Campello, Ricardo JGB and Moulavi, Davoud and Sander, J{\"o}rg},
  booktitle={Advances in Knowledge Discovery and Data Mining},
  pages={160--172},
  year={2013}
}
\end{document}


\title{{TaskTok}: Delving into Task Tokens for Task-driven Image Restoration\texorpdfstring{\\[0.25em]
{\large Supplementary Material}}{ - Supplementary Material}}

\titlerunning{TaskTok}

\author{Hongjae Lee\orcidlink{0000-0001-5312-139X} \and
Sojung Kang\orcidlink{0009-0009-6928-2900} \and
Jaeseong Yu\orcidlink{0009-0005-1427-6651} \and
Seung-Won Jung\thanks{Corresponding author.}\orcidlink{0000-0002-0319-4467}}

\authorrunning{H. Lee \etal}

\institute{Korea University\\
\email{\{jimmy9704, topazsco, jsyu624, swjung83\}@korea.ac.kr}}

\appendix
\maketitle

We organize our supplementary material as follows:
\begin{itemize}
    \item \cref{sec:latency} reports the parameter count and latency breakdown of each module in TaskTok.
    \item \cref{sec:loss} ablates loss weights and greedy-initialization sample size.
    \item \cref{sec:greedy} details the greedy token-importance ordering procedure.
    \item \cref{sec:greedy_results} presents greedy search results across downstream tasks.
    \item \cref{sec:difference_sup} provides qualitative examples of visual differences across multiple token indices.
    \item \cref{sec:word_cloud} analyzes task-specific attribute preferences.
    \item \cref{sec:imple} provides implementation details and training configurations.
    \item \cref{sec:additional_visuals} provides additional visual results.
\end{itemize}

\section{Parameter and Latency Breakdown of TaskTok-64}
\label{sec:latency}

\cref{tab:latency_breakdown} reports the parameter count and latency of each module in TaskTok-64, including pre-restoration (Pre), encoder (Enc), token refiner (Ref), and decoder (Dec). The token refinement module accounts for only a small fraction of the total cost.

\begin{table}[!ht]
\caption{Parameter and latency breakdown of TaskTok-64. Pre, Enc, Ref, and Dec denote pre-restoration, encoder, token refiner, and decoder, respectively.}
\label{tab:latency_breakdown}
\centering
\setlength{\tabcolsep}{8pt}
\renewcommand{\arraystretch}{.55}
\begin{tabular}{lccccc}
\toprule
Metric & Pre & Enc & Ref & Dec & Overall \\
\midrule
Latency (ms/img) & 5.45 & 1.83 & 0.08 & 5.90 & 13.26 \\
Params (M)       & 5.17 & 85.94 & 19.02 & 303.50 & 413.63 \\
\bottomrule
\end{tabular}
\end{table}


\begin{figure}[!t]
  \centering
  \begin{minipage}[t]{0.675\linewidth}
    \centering
    \includegraphics[width=.98\linewidth]{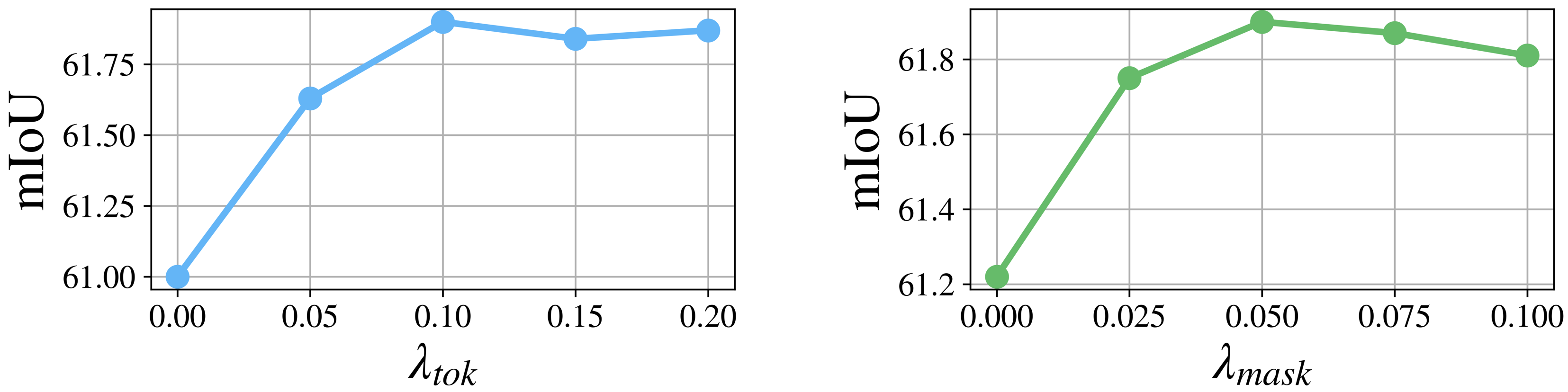}
    \caption{Sensitivity to the loss weights $\lambda_{tok}$ and $\lambda_{mask}$ under the Mix-B setting. The ablation is conducted on the segmentation task.}
    \label{fig:loss_ablation}
  \end{minipage}
  \hfill
  \begin{minipage}[t]{0.295\linewidth}
    \centering
    \includegraphics[width=.98\linewidth]{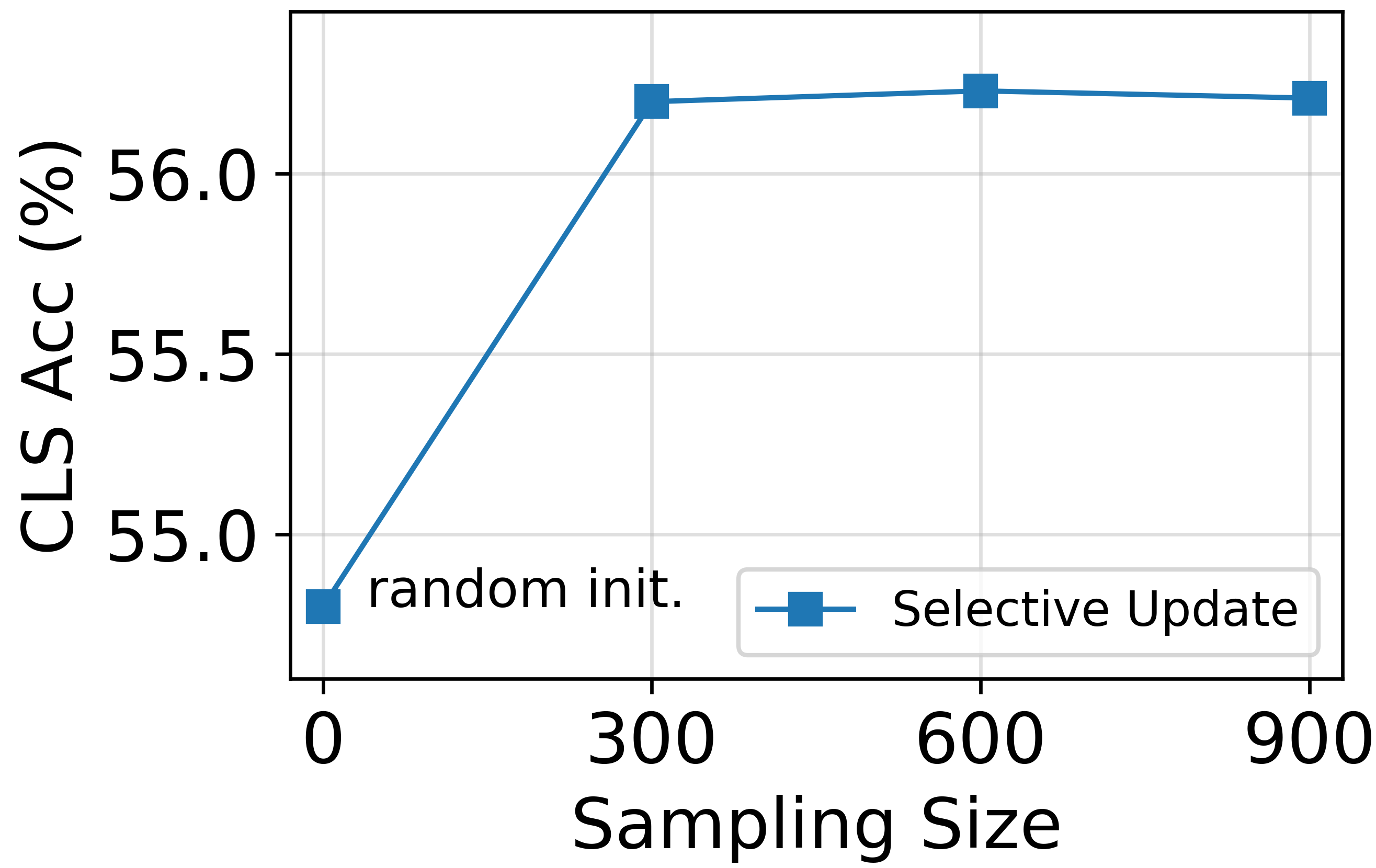}
    \caption{Greedy initialization ablation.}
    \label{fig:sampling}
  \end{minipage}
\end{figure}

\section{Loss Weight and Greedy Initialization Ablation}
\label{sec:loss}

\textbf{Loss weight ablation.} We study the sensitivity to the loss weights $\lambda_{tok}$ and $\lambda_{mask}$ under the Mix-B setting. As shown in \cref{fig:loss_ablation}, performance improves up to $\lambda_{tok}=0.1$ and $\lambda_{mask}=0.05$, and then saturates or slightly declines. We therefore adopt $(\lambda_{tok}, \lambda_{mask}) = (0.1, 0.05)$ by default.

\begin{algorithm}[!b]
\caption{Greedy Token-Importance Ordering}
\label{alg:greedy_order}
\KwIn{HQ tokens $\mathbf{Z}^{HQ}=[\mathbf{z}_1^{HQ};\dots;\mathbf{z}_K^{HQ}]$, LQ tokens $\mathbf{Z}^{LQ}=[\mathbf{z}_1^{LQ};\dots;\mathbf{z}_K^{LQ}]$, task model $\mathtt{Task}(\cdot)$, metric $\mathtt{Score}(\cdot)$}
\KwOut{Token importance rank $\boldsymbol{\pi}$}

Initialize $\mathbf{Z}\leftarrow \mathbf{Z}^{HQ}$, $\mathcal{U}\leftarrow\{1,\dots,K\}$, and $\boldsymbol{\pi}_{\text{rev}}\leftarrow [\ ]$\;

\While{$\mathcal{U}\neq\emptyset$}{
  \ForEach{$i\in\mathcal{U}$}{
    Form $\mathbf{Z}^{(i)}$ by replacing $\mathbf{z}_i \leftarrow \mathbf{z}_i^{LQ}$ in $\mathbf{Z}$\;
    $s_i \leftarrow \mathtt{Score}(\mathtt{Task}(\mathbf{Z}^{(i)}))$\;
  }
  Select $i^\ast \leftarrow \arg\max_{i\in\mathcal{U}} s_i$\ \tcp*{Identify least important token}
  Update $\mathbf{Z} \leftarrow \mathbf{Z}^{(i^\ast)}$\;
  Append $i^\ast$ to $\boldsymbol{\pi}_{\text{rev}}$\;
  Remove $i^\ast$ from $\mathcal{U}$\;
}
\Return $\boldsymbol{\pi} \leftarrow \mathrm{Reverse}(\boldsymbol{\pi}_{\text{rev}})$ \tcp*{Order: Most $\to$ Least}
\end{algorithm}

\noindent \textbf{Greedy initialization ablation.} We further evaluate the sensitivity to the number of images used for greedy initialization. Specifically, we vary the number of images from 0 to 900, where 0 corresponds to random initialization without the greedy ranking. As shown in~\cref{fig:sampling}, 300 images are sufficient for initialization, since the switch is further optimized with downstream supervision, mitigating possible sample bias.

\section{Details of Greedy Token-Importance Ordering}
\label{sec:greedy}

\cref{alg:greedy_order} details the proposed greedy search procedure for obtaining the token-importance ordering for a given downstream task. Starting from the HQ token sequence, we iteratively replace one token with its LQ counterpart and evaluate the resulting performance. At each step, the token whose replacement yields the highest performance is treated as the least task-relevant and removed. Repeating this process yields an ordering from least to most important, which is reversed to obtain the final rank $\boldsymbol{\pi}$.

\begin{figure}[!b]
  \centering
  \includegraphics[width=.98\textwidth]{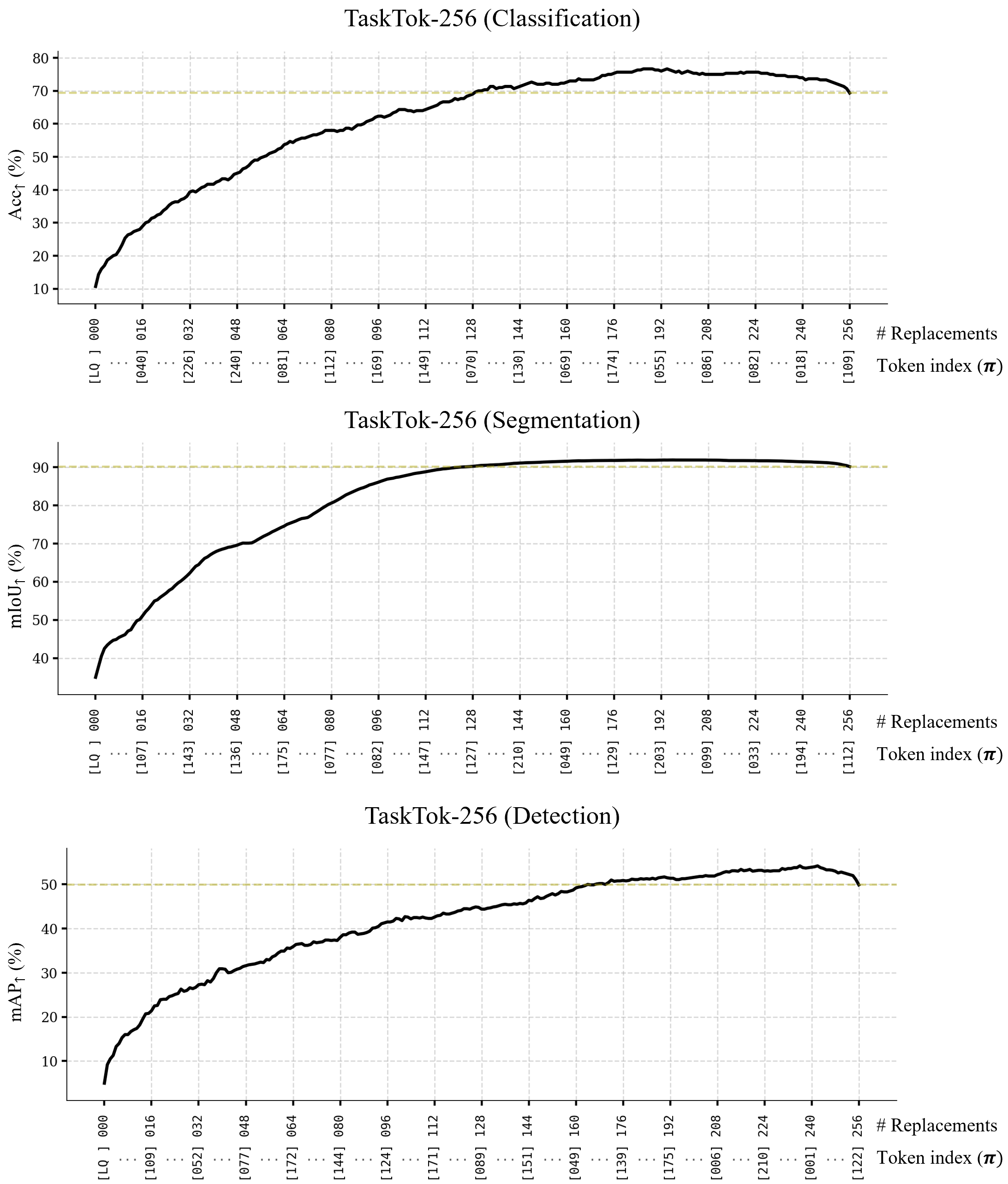}
  \caption{Greedy search results for CLS, Seg, and Det with TaskTok-256.}
  \label{fig:greedy_256}
\end{figure}

\begin{figure}[p]
  \centering
  \includegraphics[width=.98\textwidth]{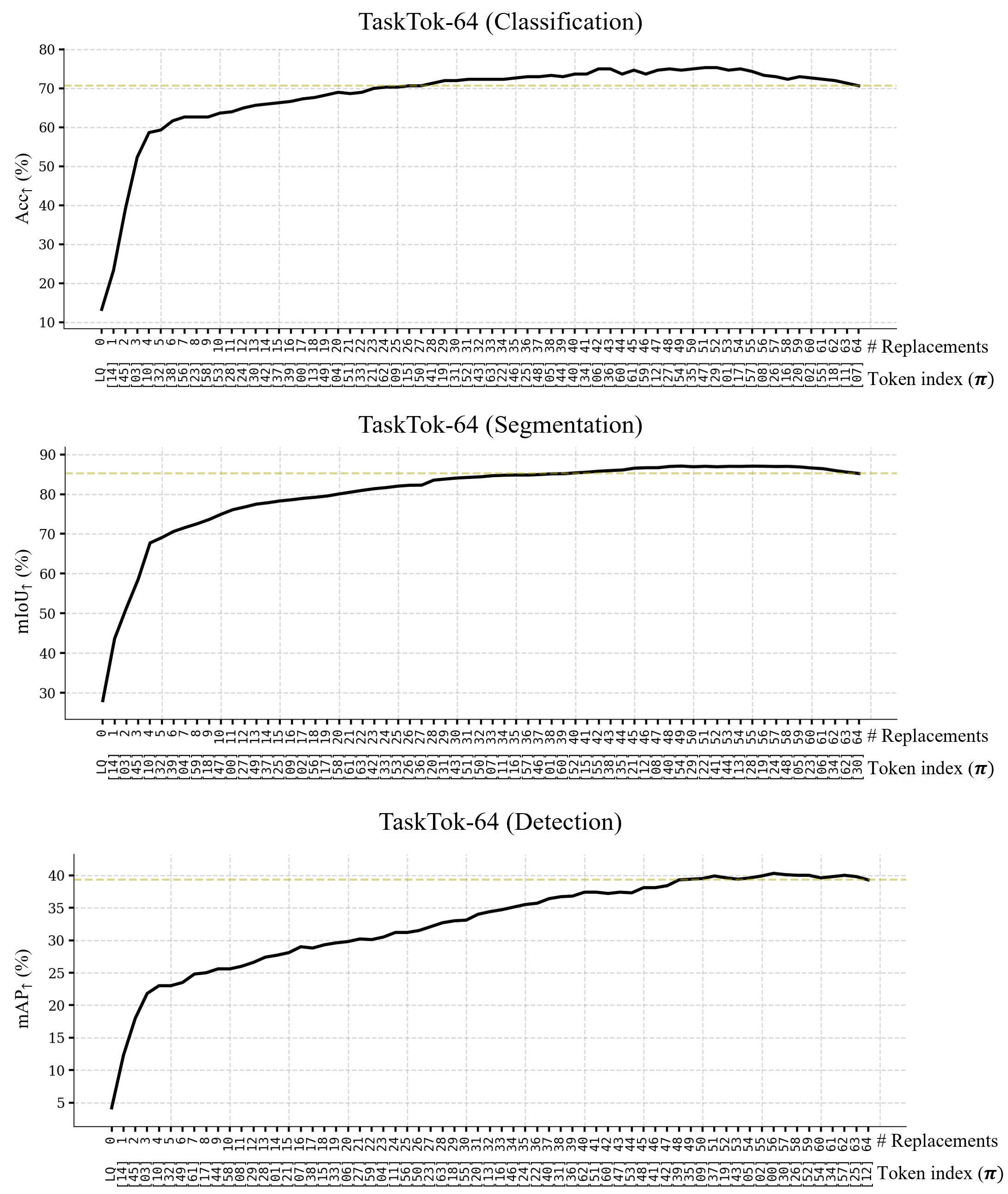}
  \caption{Greedy search results for CLS, Seg, and Det with TaskTok-64.}
  \label{fig:greedy_64}
\end{figure}

\begin{figure}[p]
  \centering
  \includegraphics[width=\textwidth]{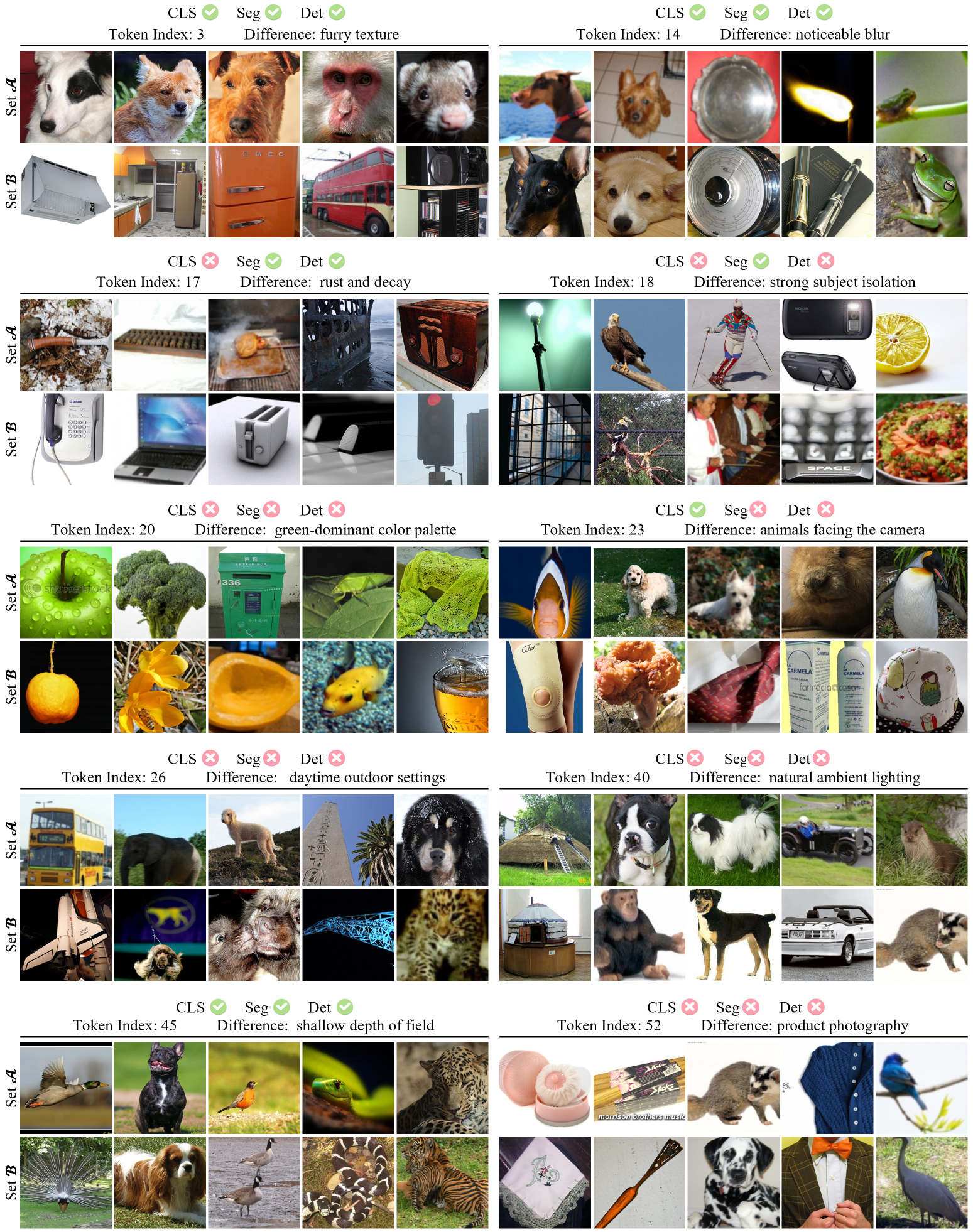}
  \caption{Qualitative examples of Set $\mathcal{A}$ and Set $\mathcal{B}$ for multiple token indices. For each token index, the two sets are drawn from the most distant clusters in the token representation space, and the accompanying label summarizes their dominant visual difference. Check and cross marks indicate whether the token is selected by CLS, Seg, and Det.}
  \label{fig:difference_sup}
\end{figure}

\begin{figure}[!h]
  \centering
  \includegraphics[width=.9\textwidth]{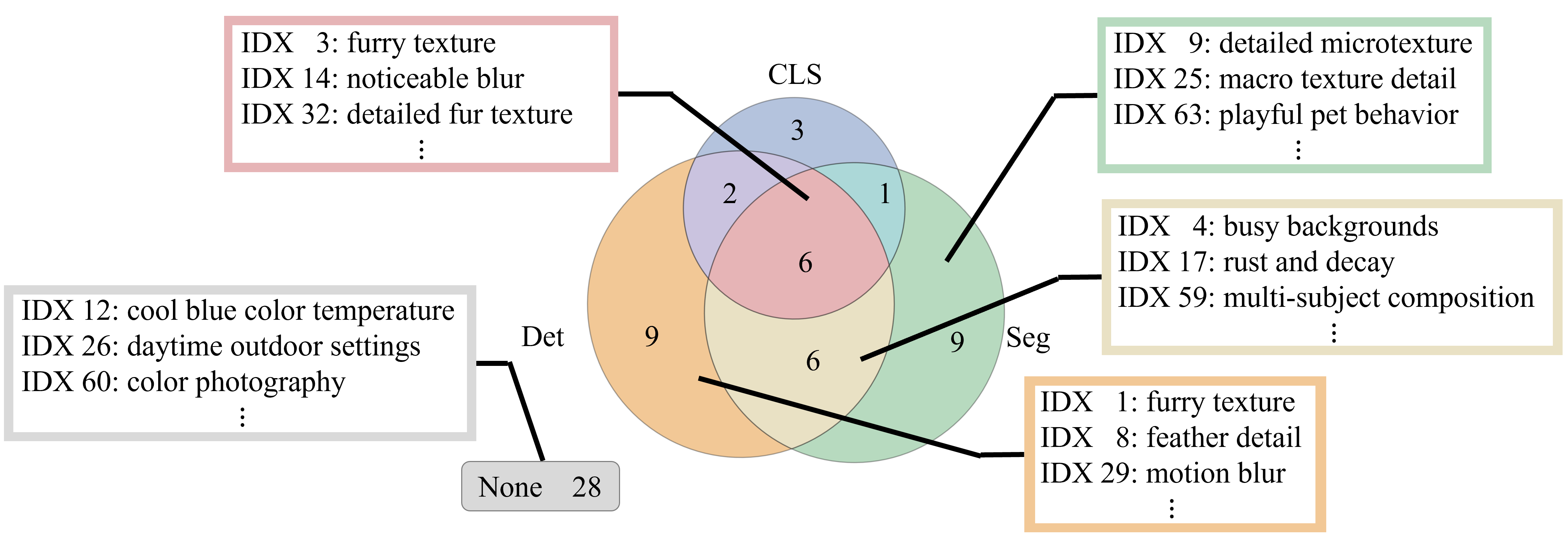}
  \caption{Overlap of task-specific token selections in TaskTok-64. Universally selected tokens across tasks capture fine-grained structural details, while consistently unselected tokens correspond to global appearance factors.}
  \label{fig:diagram}
\end{figure}

\section{Greedy Search Results Across Downstream Tasks}
\label{sec:greedy_results}

We further present the greedy search results for CLS, Seg, and Det using both TaskTok-64 and TaskTok-256. As shown in \cref{fig:greedy_256,fig:greedy_64}, restoring more tokens does not necessarily improve downstream task performance. For all three tasks, performance often saturates after a task-relevant subset is restored, and restoring additional tokens may even lead to performance degradation, suggesting semantic shift. These results show that using more tokens does not guarantee better task performance.

\section{Qualitative Visual Differences Across Token Indices}
\label{sec:difference_sup}
\cref{fig:difference_sup} provides qualitative examples of Set $\mathcal{A}$ and Set $\mathcal{B}$ for multiple token indices. For each token index, the two sets are drawn from the most distant clusters in the token representation space, and the accompanying difference caption summarizes their dominant visual difference. These examples demonstrate that our method effectively captures semantically meaningful visual variations reflected in natural language.

These visualizations also help interpret task-specific token selections. For instance, token 3 (furry texture) distinguishes furry animals from smooth objects. Because such texture cues are broadly useful across recognition tasks, this token is selected by all three tasks. Similarly, token 18 (subject isolation) separates salient objects from complex backgrounds. As figure--ground separation is crucial for dense prediction, it is exclusively selected by the segmentation task.

Conversely, tokens primarily capturing non-structural attributes are consistently unselected across all tasks. This includes variations in color (\eg, token 20: green-dominant color palette), illumination (tokens 26 and 40), and style (token 52), which lack direct relevance to the semantic content required for the downstream tasks.

\section{Task-Specific Attribute Preference Analysis}
\label{sec:word_cloud}

To better understand which visual attributes are preferred by each downstream task, \cref{fig:diagram} summarizes the overlap of selected tokens across CLS, Seg, and Det, with each subset annotated by its token count and three representative token-difference examples. The subset shared by all three tasks is primarily associated with broadly useful cues related to fine-grained detail and morphological sharpness (\eg, furry texture and noticeable blur), whereas the 28 consistently unselected tokens tend to correspond to global appearance factors such as color or illumination (\eg, cool blue color temperature and daytime outdoor settings). This suggests that such global attributes are either already sufficiently preserved in the degraded inputs or do not decisively affect downstream task performance. Meanwhile, Seg and Det tend to share tokens related to background clutter and scene complexity, consistent with their need for object localization and boundary delineation. Notably, because segmentation requires dense pixel-level precision, Seg exhibits heightened sensitivity to localized detail variations (\eg, detailed microtexture and macro texture detail). In addition, tokens selected only by Det are often associated with object-centric visual cues (\eg, furry texture and motion blur), which may be useful for distinguishing object instances from the surrounding scene. The full list of token-wise difference labels and task-specific selection results is provided in \cref{tab:token_att_sets}.

\begin{table}[p]
\centering
\caption{Token-wise attribute differences grouped by task-specific selection patterns.}
\label{tab:token_att_sets}
\scriptsize
\setlength{\tabcolsep}{2.5pt}
\renewcommand{\arraystretch}{1.1}
\begin{tabular}{c L{3.2cm} c c c @{\hspace{6pt}} c L{3.2cm} c c c}
\toprule
IDX & Difference & CLS & Seg & Det & IDX & Difference & CLS & Seg & Det \\
\midrule

\multicolumn{10}{l}{\textbf{CLS only}} \\
23 & animals facing the camera & \cmark & \xmark & \xmark &
24 & busy backgrounds & \cmark & \xmark & \xmark \\
53 & camouflage patterns & \cmark & \xmark & \xmark &
\multicolumn{5}{l}{} \\
\midrule

\multicolumn{10}{l}{\textbf{Seg only}} \\
0 & busy, cluttered scenes & \xmark & \cmark & \xmark &
2 & low image noise & \xmark & \cmark & \xmark \\
9 & detailed microtexture & \xmark & \cmark & \xmark &
18 & strong subject isolation & \xmark & \cmark & \xmark \\
25 & macro texture detail & \xmark & \cmark & \xmark &
37 & close-up framing & \xmark & \cmark & \xmark \\
39 & pets on furniture & \xmark & \cmark & \xmark &
47 & household objects & \xmark & \cmark & \xmark \\
63 & playful pet behavior & \xmark & \cmark & \xmark &
\multicolumn{5}{l}{} \\
\midrule

\multicolumn{10}{l}{\textbf{Det only}} \\
1 & furry texture & \xmark & \xmark & \cmark &
6 & silhouette against sky & \xmark & \xmark & \cmark \\
7 & clear blue sky & \xmark & \xmark & \cmark &
8 & feather detail & \xmark & \xmark & \cmark \\
15 & tight close-up framing & \xmark & \xmark & \cmark &
21 & low-key dramatic lighting & \xmark & \xmark & \cmark \\
29 & motion blur & \xmark & \xmark & \cmark &
33 & deep depth of field & \xmark & \xmark & \cmark \\
44 & blue-dominant color palette & \xmark & \xmark & \cmark &
\multicolumn{5}{l}{} \\
\midrule

\multicolumn{10}{l}{\textbf{CLS \& Seg}} \\
56 & ground-level perspective & \cmark & \cmark & \xmark &
\multicolumn{5}{l}{} \\
\midrule

\multicolumn{10}{l}{\textbf{CLS \& Det}} \\
28 & isolated subject on plain background & \cmark & \xmark & \cmark &
38 & product photography & \cmark & \xmark & \cmark \\
\midrule

\multicolumn{10}{l}{\textbf{Seg \& Det}} \\
4 & busy backgrounds & \xmark & \cmark & \cmark &
17 & rust and decay & \xmark & \cmark & \cmark \\
27 & jewelry product photography & \xmark & \cmark & \cmark &
49 & snowy high-key scenes & \xmark & \cmark & \cmark \\
59 & multi-subject composition & \xmark & \cmark & \cmark &
61 & vintage aesthetic & \xmark & \cmark & \cmark \\
\midrule

\multicolumn{10}{l}{\textbf{CLS \& Seg \& Det}} \\
3 & furry texture & \cmark & \cmark & \cmark &
10 & soft focus blur & \cmark & \cmark & \cmark \\
14 & noticeable blur & \cmark & \cmark & \cmark &
32 & detailed fur texture & \cmark & \cmark & \cmark \\
45 & shallow depth of field & \cmark & \cmark & \cmark &
58 & high sharpness on main subject & \cmark & \cmark & \cmark \\
\midrule

\multicolumn{10}{l}{\textbf{None}} \\
5 & solid white background & \xmark & \xmark & \xmark &
11 & low-key lighting & \xmark & \xmark & \xmark \\
12 & cool blue color temperature & \xmark & \xmark & \xmark &
13 & soft shadows & \xmark & \xmark & \xmark \\
16 & studio product photography & \xmark & \xmark & \xmark &
19 & centered framing & \xmark & \xmark & \xmark \\
20 & green-dominant color palette & \xmark & \xmark & \xmark &
22 & cool blue color palette & \xmark & \xmark & \xmark \\
26 & daytime outdoor settings & \xmark & \xmark & \xmark &
30 & warm earthy color palette & \xmark & \xmark & \xmark \\
31 & black-and-white & \xmark & \xmark & \xmark &
34 & solid white background & \xmark & \xmark & \xmark \\
35 & warm sunset lighting & \xmark & \xmark & \xmark &
36 & vintage antiques & \xmark & \xmark & \xmark \\
40 & natural ambient lighting & \xmark & \xmark & \xmark &
41 & product photography & \xmark & \xmark & \xmark \\
42 & museum exhibits & \xmark & \xmark & \xmark &
43 & vintage film aesthetic & \xmark & \xmark & \xmark \\
46 & uneven exposure & \xmark & \xmark & \xmark &
48 & retro technology & \xmark & \xmark & \xmark \\
50 & earthy color palette & \xmark & \xmark & \xmark &
51 & catalog-style product shot & \xmark & \xmark & \xmark \\
52 & product photography & \xmark & \xmark & \xmark &
54 & aerial vehicles & \xmark & \xmark & \xmark \\
55 & smooth sky gradients & \xmark & \xmark & \xmark &
57 & fashion accessories & \xmark & \xmark & \xmark \\
60 & color photography & \xmark & \xmark & \xmark &
62 & saturated colors & \xmark & \xmark & \xmark \\
\bottomrule
\end{tabular}
\end{table}

\section{Additional Implementation Details}
\label{sec:imple}

TaskTok uses a lightweight SwinIR~\cite{swinir} as the pre-restoration module, consisting of 12 Swin transformer blocks. Following prior work~\cite{edtr}, we initialize the pre-restorer with ImageNet-pretrained weights and keep it frozen during training. TaskTok-64 and TaskTok-256 adopt the resolution-256 TiTok-BL-64 and TiTok-SL-256 VQ tokenizers~\cite{titok}, respectively. For fair comparison, both TaskTok and the TiTok baselines are trained using the same two-stage training protocol and objective functions as EDTR~\cite{edtr}. In both cases, the TiTok encoder is frozen throughout training.

\section{Additional Qualitative Results}
\label{sec:additional_visuals}
\begin{figure}[!h]
  \centering
  \includegraphics[width=.98\textwidth]{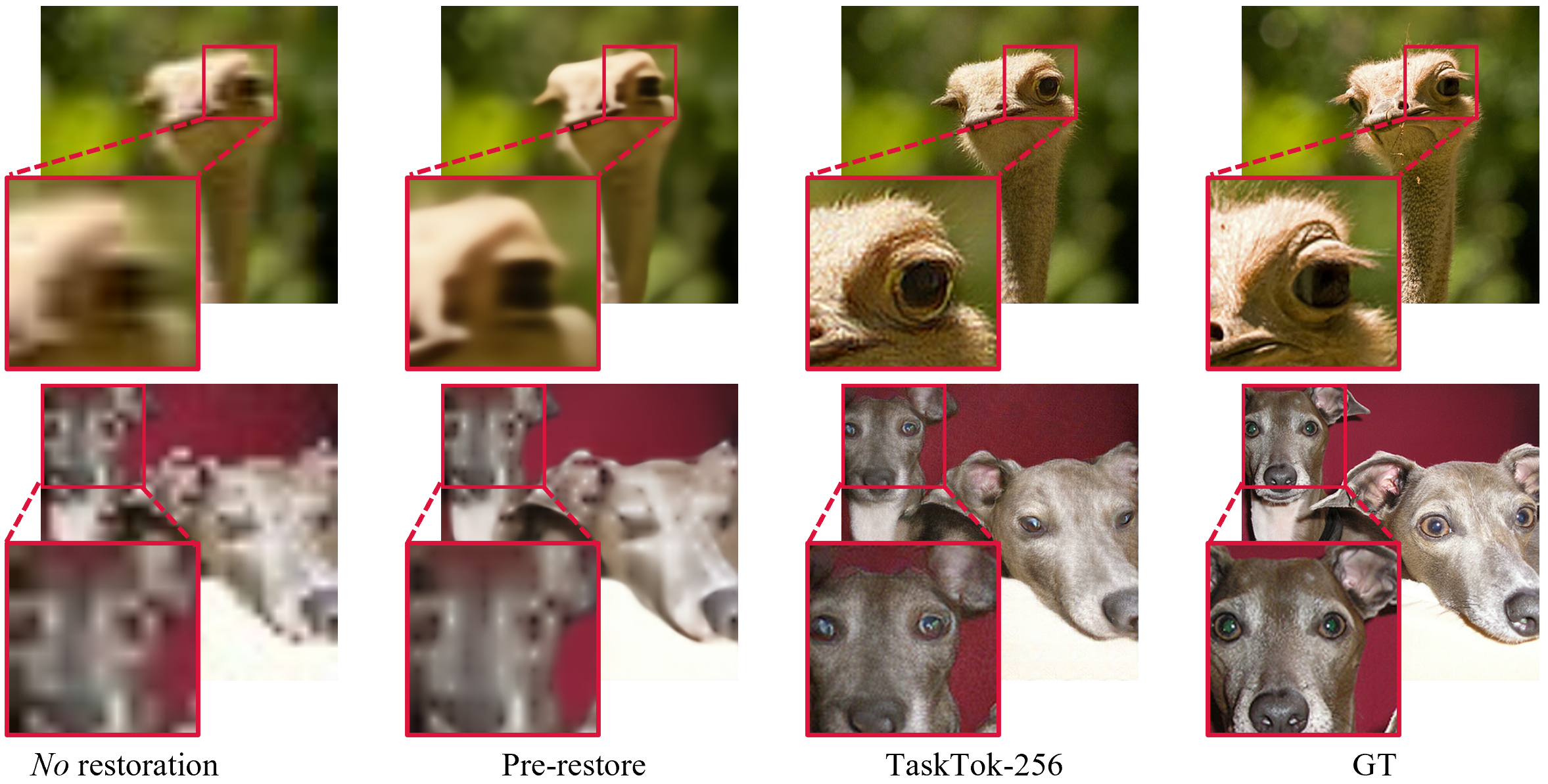}
  \caption{Qualitative examples of the pre-restoration outputs and the TaskTok results.}
  \label{fig:pre_tasktok}
\end{figure}

\begin{figure}[!b]
  \centering
  \includegraphics[width=.9\textwidth]{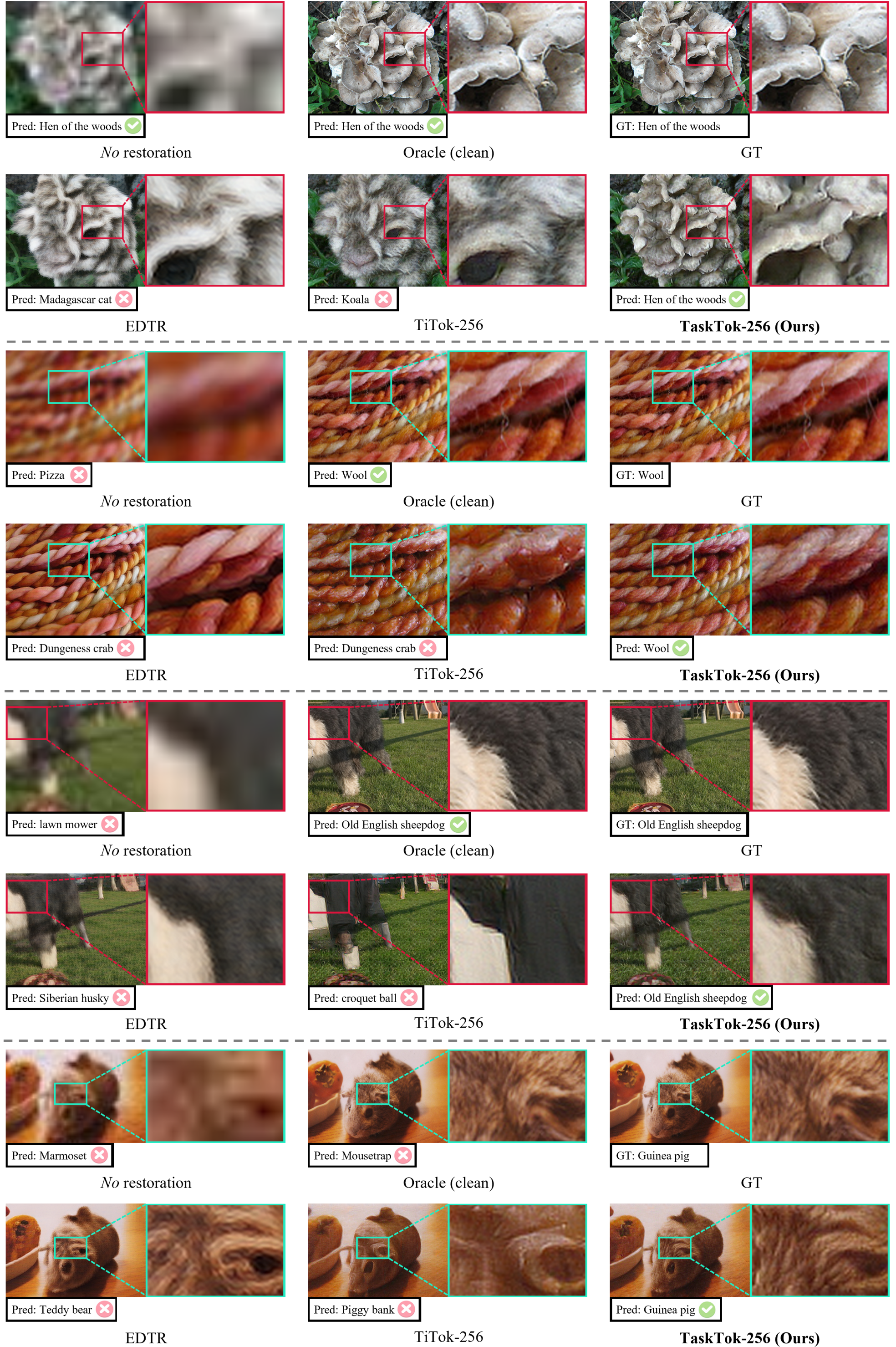}
  \caption{Additional qualitative results for image classification.}
  \label{fig:sup_cls}
\end{figure}

\begin{figure}[p]
  \centering
  \includegraphics[width=.9\textwidth]{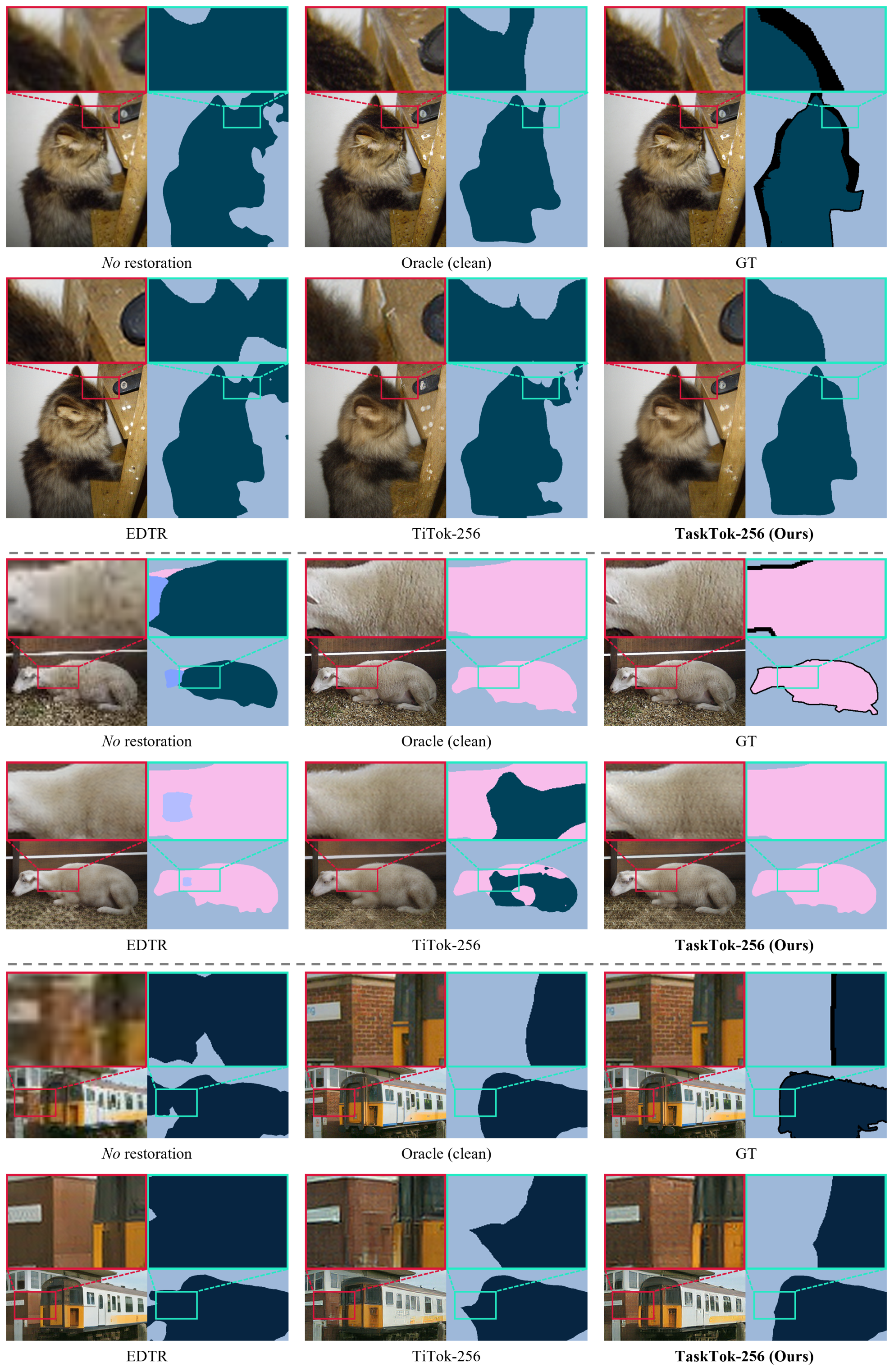}
  \caption{Additional qualitative results for semantic segmentation.}
  \label{fig:sup_seg}
\end{figure}

\begin{figure}[p]
  \centering
  \includegraphics[width=.9\textwidth]{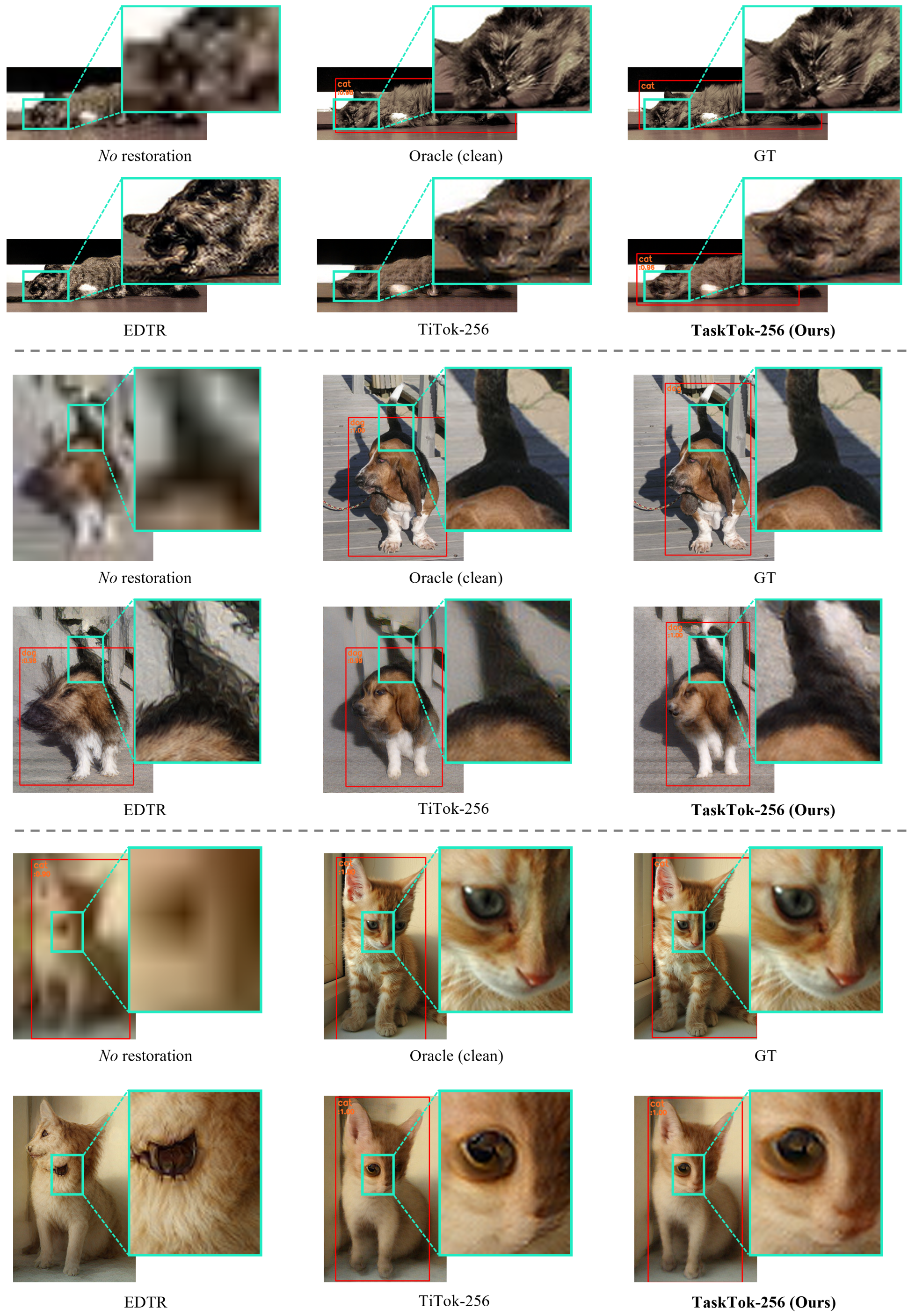}
  \caption{Additional qualitative results for object detection.}
  \label{fig:sup_det}
\end{figure}

Additional qualitative results are presented in \cref{fig:pre_tasktok,fig:sup_cls,fig:sup_seg,fig:sup_det}.

\clearpage

%
%
\bibliographystyle{splncs04}
\bibliography{main}